%% file: main.tex
\documentclass[5p,times,twocolumn,sort,number,preprint,numafflabel]{elsarticle} %

\usepackage{amssymb}
\usepackage{amsmath}

\usepackage[switch]{lineno} 

\usepackage[normalem]{ulem}

\usepackage{xcolor}
\usepackage{amsmath}
\usepackage[table]{xcolor} 
\usepackage{amssymb}
\usepackage{comment}
\usepackage{booktabs}
\usepackage[colorlinks=true,
            linkcolor=blue,
            citecolor=blue,
            urlcolor=blue]{hyperref}

\newcommand\NOTE[3]{\textcolor{#1}{[#2: #3]}}
\definecolor{jorange}{rgb}{8,.5,.0}
\newcommand\linus[1]{\NOTE{jorange}{Linus}{#1}}
\definecolor{bluegreen}{rgb}{.05,.6,.7}
\newcommand\svenja[1]{\NOTE{bluegreen}{Svenja}{#1}}
\newcommand\marc[1]{\NOTE{magenta}{Marc}{#1}}
\newcommand\matthias[1]{\NOTE{olive}{Matthias}{#1}}
\newcommand\bernhard[1]{\NOTE{blue}{Bernhard}{#1}}

\definecolor{darkgreen}{rgb}{0,.6,.0}

\renewcommand\marc[1]{}
\renewcommand\linus[1]{}
\renewcommand\svenja[1]{}
\renewcommand\matthias[1]{}
\renewcommand\bernhard[1]{}

\newcommand\old[1]{}
\newcommand\new[1]{{#1}}

\journal{Computers \& Graphics}

\begin{document}

\author[label1,label2]{Svenja Strobel}
\author[label2]{Matthias Innmann}
\author[label1]{Bernhard Egger}
\author[label1]{Marc Stamminger}
\author[label4,label1,label3]{Linus Franke\corref{cor1}}
\affiliation[label1]{organization={Friedrich-Alexander-Universität Erlangen-Nürnberg},
            addressline={Cauerstrasse 11},
            city={Erlangen},
            postcode={91058},
            state={Bavaria},
            country={Germany}}
\affiliation[label2]{organization={NavVis GmbH},
            addressline={Blutenburgstraße 18},
            city={München},
            postcode={80636},
            state={Bavaria},
            country={Germany}}
\affiliation[label3]{organization={Inria, Université Côte d'Azur},
            addressline={2004 Rte des Lucioles},
            city={Valbonne},
            postcode={06560},
            country={France}}
\affiliation[label4]{organization={Julius-Maximilians-Universität Würzburg},
            addressline={Am Hubland},
            city={Würzburg},
            postcode={97074},
            state={Bavaria},
            country={Germany}}
\cortext[cor1]{Corresponding author. Email: \{firstname.lastname\}@uni-wuerzburg.de}

\title{SurfFill: Completion of LiDAR Point Clouds via Gaussian Surfel Splatting}

\input{00-abstract.tex}
\maketitle

\begin{figure*}[!t]
\vspace{-2mm}
\includegraphics[width=.99\linewidth]{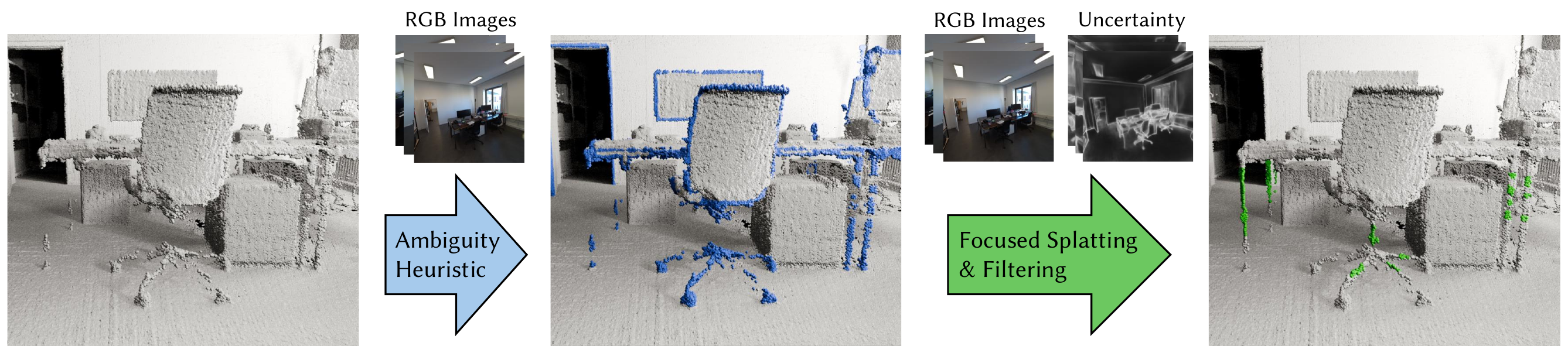}%
\vspace{-2mm}
\caption[]{For a given LiDAR scan, we complete the point cloud by carefully integrating Gaussian surfel-based 3D reconstruction.
We analyze erroneous regions and identify {\em ambiguity} in the point cloud as well as areas of high {\em uncertainty} in captured visual data. Missing structures are reconstructed with a {\em focused 2D Gaussian Splatting} technique, followed by filtering and sampling for point cloud completion. We extend this with a divide-and-conquer scheme, allowing us to process this building-scale point cloud in less than 75 minutes.}%
\label{fig:teaser}%
\vspace{-1mm}
\end{figure*}

\input{00-abstract.tex}
\input{01-intro.tex}

\input{02-related.tex}
\input{03-main.tex}
\input{04-eval.tex}

\input{05-conc.tex}

\section*{Acknowledgements}
Our gratitude goes to Stefan Romberg, Michael Gerstmayr, and Tim Habigt for the productive discussions. 
The project was supported by the Bayerische Forschungsstiftung (Bavarian Research Foundation) AZ-1626-24.
Linus Franke was supported in part by the ERC Advanced Grant NERPHYS (101141721, {https://project.inria.fr/nerphys}).
Views and opinions expressed are however those of the authors only and do not necessarily reflect those of the EU or the European Research Council. Neither the EU nor the granting authority can be held responsible for them.
The authors gratefully acknowledge the scientific support and HPC resources provided by the Erlangen National High Performance Computing Center (NHR@FAU) of the Friedrich-Alexander-Universität Erlangen-Nürnberg (FAU) under the NHR project \textit{b162dc}. NHR funding is provided by federal and Bavarian state authorities. NHR@FAU hardware is partially funded by the German Research Foundation (DFG) – 440719683.

{
\small
\bibliographystyle{elsarticle-num}
\bibliography{egbib}      
\clearpage
}

\appendix
\input{99-suppl.tex}

\end{document}

%% file: 00-abstract.tex
\begin{abstract}
LiDAR-captured point clouds are often considered the gold standard in active 3D reconstruction.
While their accuracy is exceptional in flat regions, the capturing is susceptible to missing small geometric structures, thin edges, and structures exhibiting challenging surface properties.
Alternatively, capturing multiple photos of the scene and applying 3D photogrammetry can infer these details as they often represent feature-rich regions.
However, the accuracy of LiDAR for featureless regions is rarely reached.

Therefore, we suggest combining the strengths of LiDAR and camera-based capture by introducing SurfFill: a Gaussian surfel-based LiDAR completion scheme.
We analyze LiDAR capturings and attribute LiDAR beam divergence as a main factor for artifacts, manifesting mostly at thin structures and edges.
We use this insight to introduce an \textit{ambiguity heuristic} for completed scans by evaluating the change in density in the point cloud.
This allows us to identify points close to missed areas, which we can then use to grow additional points from to complete the scan.
For this point growing, we employ Gaussian surfels and focus optimization and densification on these ambiguous areas.
Finally, Gaussian primitives of the reconstruction in ambiguous areas are extracted and sampled for points to complete the point cloud.
To address the challenges of large-scale reconstruction, we extend this pipeline with a divide-and-conquer scheme for building-sized point cloud completion.
We evaluate on the task of LiDAR point cloud completion of synthetic and real-world scenes and find that our method outperforms previous reconstruction methods.

\vspace{-21.5em}
\centering \url{https://lfranke.github.io/surffill}
\vspace{20.5em}
\end{abstract}

%% file: 01-intro.tex
\section{Introduction}

Light-based sensing via light detection and ranging (LiDAR) works by scanning the environment with rotating bundles of laser beams and measuring their time of flight.
It has been the gold standard for environment and object scanning and is used as ground truth for benchmarks in many tasks related to 3D data~\cite{Knapitsch2017,yeshwanthliu2023scannetpp,Liao2022kitti}.
Due to the LiDAR's panoramic 360-degree field of view, LiDAR SLAM methods generate models with high global consistency and precision, even for outdoor scenes~\cite{hong2023comparison}.
However, \new{because each laser beam has a finite cross-section that grows with distance (\textit{beam divergence}), it can simultaneously illuminate foreground and background surfaces near geometric discontinuities. The sensor then records a \textit{mixed pixel}, a spurious range measurement that is neither surface, and subsequent pre-processing filters discard these ambiguous returns as outliers.} The result is that LiDAR point clouds exhibit systematic gaps precisely at thin structures, silhouettes, and edges: the locations where beam divergence causes the footprint to straddle multiple depths.
Solving these failure cases may be expensive, requiring additional scans, or even manual fixing by a 3D artist.
Furthermore, while errors manifest as partial, incomplete shapes, the sheer scale and resource demand of these datasets renders typical shape completion methods~\cite{zhuang2024survey} unfeasible.

In contrast, photos of the scene are inexpensive to acquire and allow common 3D reconstruction methods to match image-to-image patches for geometric reconstruction~\cite{schoenberger2016mvs}.
Such Multi-view Stereo (MVS) reconstructions can be very detailed, but they often exhibit holes in feature-sparse areas.
As an alternative, inverse rendering methods, e.g.\ NeuS~\cite{wang2023neus2} and Neuralangelo~\cite{li2023neuralangelo} using neural implicit representations or 2D Gaussian Splatting (2DGS)~\cite{huang20242d} using Gaussian surfels, reconstruct finer geometry.
However, they also suffer from artifacts in feature-sparse areas.
While both directions do not reach the global precision of LiDAR scans, they perform best in feature-rich regions, e.g.\ along edges which are difficult for LiDAR capture due to beam divergence.

A key insight is that LiDAR scans and image-based reconstruction thus complement each other very well, as seen in the results in Fig.~\ref{fig:teaser}.
Consequently, this paper aims to leverage the strengths of both modalities by combining detailed LiDAR scanning in flat, feature-sparse areas with photometric 3D reconstruction along silhouettes and edges.

The key problem is how to integrate photometric reconstruction with completed LiDAR scans, as we need to \textit{identify} areas where the LiDAR scans are erroneous and simultaneously perform a \textit{focused} reconstruction to avoid adding outliers or low-accuracy samples from the photometric reconstruction to the point cloud.

Directly identifying missing scanned areas is difficult, as, for example, a four-legged table missing one leg (seen in Fig.~\ref{fig:teaser} (left)) is indistinguishable from a three-legged table from the point cloud alone.
However, \new{beam divergence leaves a characteristic fingerprint: rather than a clean boundary between present and absent geometry, pre-processing filters produce a sparse \textit{transition region} of surviving but unreliable points around every gap. We exploit this by identifying \textit{ambiguous areas} (Fig.~\ref{fig:teaser} (middle), in blue) via a local point-density analysis: the thinned transition zone is the direct observable consequence of mixed-pixel removal. We introduce a technique for this, which we call the \textit{ambiguity heuristic}.}

\new{These candidate regions serve as the spatial focus for our Gaussian surfel reconstruction: their points initialize the Gaussian model in locations likely adjacent to gaps, and the optimization is driven to grow new Gaussian primitives outward from them into missing areas based on photometric error.
To further sharpen this focus, we derive a complementary signal from the captured 2D RGB images.
Specifically, we estimate monocular surface normals and infer an \textit{uncertainty} map from their spatial differences: large normal gradients indicate geometric boundaries where LiDAR data is more likely to be incomplete.
This image-space uncertainty acts as a second cue that, together with the 3D ambiguity score, jointly guides the optimization: Gaussians are encouraged to grow into regions flagged as uncertain by both modalities, while well-scanned areas are left intact.}
Furthermore, we introduce a set of losses and regularizers for this focused task and integrate a filtering and sampling scheme for acquiring points from the Gaussian model.

Although our optimization is focused on critical regions only, the overall size of LiDAR captured scenes can be massive and may overshoot common hardware limits.
We thus extend our scheme to handle this by subdividing the scene into chunks and performing a per-chunk optimization, allowing us to optimize building-sized scans with tens of millions of LiDAR points.

In summary, our contributions are as follows:
\begin{itemize}
    \item \new{An analysis of LiDAR beam divergence as the root cause of systematic gaps, and a density-based \textit{ambiguity heuristic} that exploits the characteristic transition regions it produces to identify candidate completion sites.}
    \item An integration of this heuristic into Gaussian surfel splatting for fine-detailed and focused reconstruction integrated with a filtering and sampling pipeline for new points.
    \item A set of constraints for Gaussian surfel splatting for the reconstruction of edges and exploration of space.
    \item Application of a divide-and-conquer technique to complete large-scale LiDAR datasets.
\end{itemize}

Our evaluation shows that our method produces point completion with high accuracy, outperforming common reconstruction techniques as well as shape completion techniques. Furthermore, it is able to function and complete efficiently on large-scale scenes. Project page and source code are available under: \url{https://lfranke.github.io/surffill}

%% file: 02-related.tex
\section{Related Work}\label{sec:related}

\begin{figure*}
    \centering%
    \includegraphics[width=.7\linewidth]{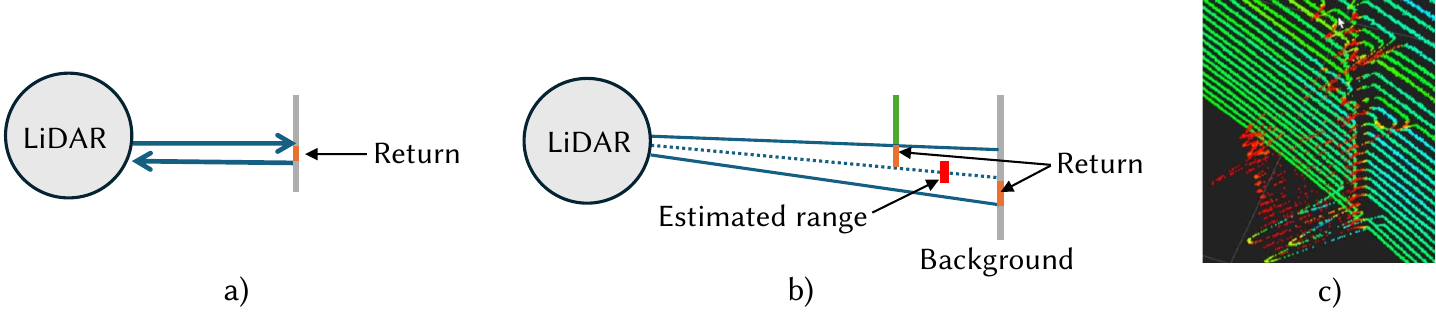}%
    \vspace{-2mm}
    \caption{High-level functionality of LiDAR and formation of mixed pixels: a)~General setup: The sensor emits and receives a light ray and calculates the range via the light travel time, b)~Formation of mixed pixels: The laser beam diverges during traveling and thus hits both foreground and background, resulting in a mixed range estimate, c)~Raw LiDAR data example with mixed pixels shown in red.
    }%
    \label{fig:lidar_arti}%
\end{figure*}

\newcommand{\relworksubsection}[1]{\paragraph*{#1}}

\relworksubsection{Shape Completion}
Direct completion techniques on point clouds have gained prominence, with recent deep learning approaches demonstrating impressive results~\cite{fei2022comprehensive}, relying on per-point features~\cite{qi2017pointnet}.
For instance, Yu et al.~\cite{yu2018pu} proposed a feature embedding and hierarchical expansion scheme to upsample point clouds, a concept further refined with GANs~\cite{li2019pu} and transformers~\cite{yu2021pointr}. 
Current popular point cloud completion methods such as \textit{PointAttN}~\cite{pointattn} and \textit{SnowflakeNet}~\cite{xiang2021snowflakenet} mainly revolve around the design of an encoder-decoder architecture for point cloud completion.
The challenge with these methods is their focus on small scales, like a few thousand points in object-centric scenes such as Shape-Net~\cite{chang2015shapenet} or isolated LiDAR sweeps~\cite{xu2019depth,xiong2023learning} and can struggle to generalize to new, varied point cloud data. 
Conversely, our approach targets fixing room and building-sized scans post-capture, handling millions of LiDAR points without specifics on scanning sweeps.

\relworksubsection{Photometric 3D Reconstruction}
Photometric 3D reconstruction has been popularized with the advent of Multi-view Stereo (MVS)~\cite{hartley2003multiple, schoenberger2016mvs, goesele2007multi}.
Following a camera calibration via Structure-from-Motion (SfM)~\cite{snavely2006photo,schonberger2016structure}, image patches in MVS are processed with plane-sweeping~\cite{collins1996space} or matched for depth along epipolar lines to maximize similarity. Recent learning-based methods extend this concept~\cite{yao2018mvsnet,xu2019multi,xu2022multi,cao2024mvsformer++, Cao2022MVSFormerMS}.
This sparked a new direction in which coarse-to-fine strategies~\cite{xu2019multi}, plane priors~\cite{xu2022multi} or vision transformers~\cite{cao2024mvsformer++, Cao2022MVSFormerMS} further improved reconstruction. %
However, texture-scarce regions often result in sparse depth maps, causing  missing structures or oversmoothed areas in fused point clouds. 
The commonly used framework \textit{COLMAP}~\cite{schonberger2016structure,schoenberger2016mvs} provides good results in a wide set of scenarios, even though not achieving the accuracy of \mbox{LiDAR} scans.

\relworksubsection{Radiance Fields and Gaussian Splatting}
MVS has traditionally served as the input for Novel View Synthesis (NVS) approaches in image-based rendering techniques~\cite{shum2000review}. More recently, however, methods have begun operating directly on point clouds, using neural point descriptors~\cite{aliev2020neural} that are optimized in a preprocessing step.
Prior work in this domain has considered LiDAR point clouds as input~\cite{ruckert2022adop,franke2024trips}; however, these approaches do not explicitly address point cloud completion.
In contrast, several methods targeting completion~\cite{xu2022point,zuo2022view,franke2023vet} are primarily designed to correct NVS artifacts and do not strive for LiDAR-grade point cloud accuracy.

Omitting the need for a complete MVS reconstruction, Neural Radiance Fields (NeRFs)~\cite{mildenhall2021nerf} reconstruct the scene as a volume, abstracted in a Multilayer Perceptron (MLP).
For geometric reconstruction, methods building on this principle~\cite{Oechsle2021ICCV, wang2021neus, yariv2021volume} simplify the reconstruction pipeline by optimizing an implicit surface representation via volume rendering. 
These methods have been extended to large-scale reconstructions through additional regularization~\cite{Yu2022MonoSDF, li2023neuralangelo, Yu2022SDFStudio, yariv2023bakedsdf} and efficient object reconstruction~\cite{wang2023neus2}.
Large-scale and LiDAR-accuracy reconstructions, however, are still challenging.
The recent state of the art is \textit{Neuralangelo}~\cite{li2023neuralangelo}, which has computation times of about a day for reconstructions.
In contrast to that, our proposed method completes similar-sized scenes in less than an hour.

To exploit the strong rasterization capabilities of recent GPUs, Kerbl and Kopanas et al.~\cite{kerbl20233d} introduced \textit{3D Gaussian Splatting (3DGS)}, where the scene is represented by anisotropic Gaussians as differentiable rendering primitives.
For optimization, they introduce and exploit an {Adaptive Density Control} module, which adds and removes Gaussians based on gradient heuristics. This allows them to efficiently reconstruct scenes from sparse starting points.
3DGS promoted a new field~\cite{wu2024recent} and has been extended regarding performance~\cite{radl2024stopthepop,hahlbohm2024htgs,franke2024vrsplatting,yu2024mip,kheradmand20243d,niemeyer2024radsplat}, compression~\cite{3DGSzip2024,Niedermayr_2024_CVPR}, scalability~\cite{kerbl2024hierarchical,lin2024vastgaussian} and dynamic scenes~\cite{luiten2023dynamic, wu20244dgs}.

Gaussian Splatting with LiDAR as structural input or for SLAM pipelines~\cite{lang2024gaussian} has also been explored~\cite{hess2024splatad,cui2024letsgo,zhou2024drivinggaussian,hwang2024vegs}, however, without targeting geometric completion.

\relworksubsection{Gaussian-based Surface Reconstruction}
In the realm of geometric reconstruction, recent works~\cite{chen2023neusg, Yu2024GSDF} integrate 3D Gaussians with neural implicit surfaces or reconstruct the scene via Poisson surface reconstruction of rendered depth maps~\cite{guedon2024sugar,dai2024high}.
\textit{Gaussian Opacity Fields (GOF)}~\cite{yu2024gaussian} utilizing 3D Gaussians, leverage explicit ray-splat intersections to enhance geometric reconstruction and incorporate Gaussian opacity in tetrahedral mesh extraction.

Huang et al.~\cite{huang20242d} introduce the use of Gaussian surfels for surface reconstruction, on which we build upon.
With \textit{2D Gaussian Splatting (2DGS)}, they employ flat primitives with optimizable central points $\mathbf{p}$, tangential vectors $\mathbf{t}_u$ and $\mathbf{t}_v$ and scaling values $\mathit{s_u}$ and $\mathit{s_v}$.
When projected, this results in an intersection point $(u,v)$.
The Gaussian's contribution is then calculated with 
$\mathcal{G}(u,v) = \exp\left(-0.5(u^2+v^2)\right).$
Surfels additionally have an optimizable opacity $\alpha$ as well as view-dependent appearance $c$ via three spherical harmonics bands.
Rendering is done via alpha-blending in an efficient tile-based renderer.
For geometric accuracy, normal consistency and depth distortion regularization are employed, and meshes are extracted via TSDF fusion.
We also use Gaussian surfels in our pipeline. However, as also noted by recent works~\cite{yu2024gaussian}, using 2DGS for fine detail reconstruction is challenging.
We solve this with a constrained reconstruction scheme in combination with scale regularization.
\new{Adjacent to our work, \textit{Li-GS}~\cite{jiang2024li} uses LiDAR depth as a surface regularizer for geometric reconstruction but explicitly restricts Gaussian primitives to the original LiDAR surface, retaining rather than filling geometric gaps. 
\textit{Tclc-GS}~\cite{zhao2024tclc} tightly couples LiDAR and camera data for real-time novel view synthesis in outdoor autonomous-driving scenarios with sparse, sweep-based acquisition patterns, a fundamentally different goal and setting from the room- and building-scale indoor completion we target.
Both methods are discussed further in Sec.~\ref{sec:evaluation} in the context of the LiDAR completion task.}

In this work, we present the first approach to evaluate artifacts and complete LiDAR scans using Gaussian Splatting.

%% file: 03-main.tex
\newcommand{\mainparagraph}[1]{\paragraph*{#1}}

\section{LiDAR Artifacts \& Incompleteness in Scans}
\label{sec:lidar_review}

LiDAR devices send a pulse of light, usually infrared, receive the reflection, and compute the range by measuring the time delta.
For an in-depth discussion of LiDARs, we refer to McManamon~\cite{mcmanamon2019}.
In this section, we explore the origin of artifacts from unfavorable geometric constellations and from non-diffuse surface reflectance in scans to arrive at a heuristic for identifying problematic regions.

\begin{figure*}
    \centering%
    \includegraphics[width=.8\linewidth]{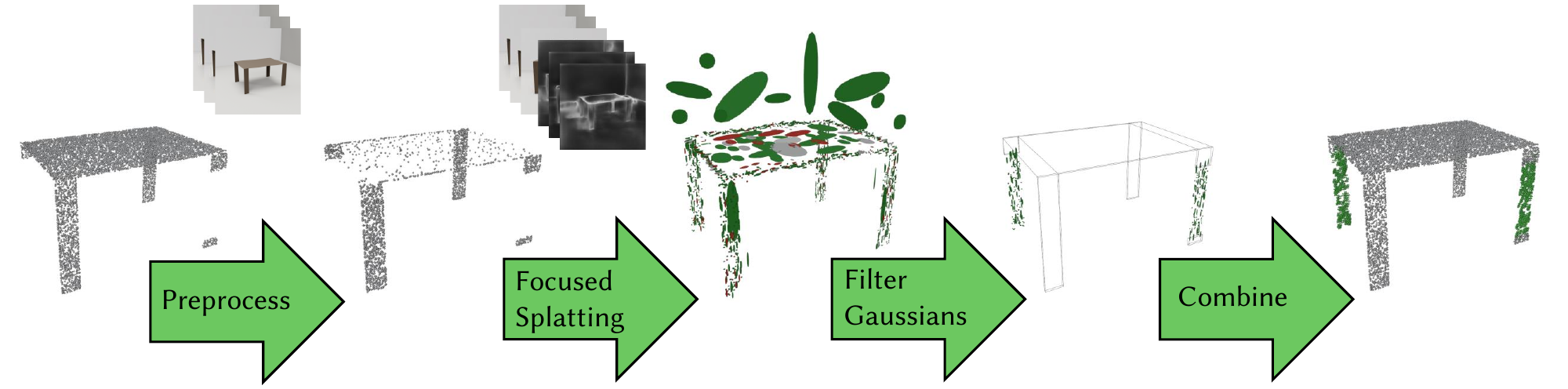}%
    \vspace{-1mm}
    \caption{Our pipeline. We estimate ambiguity and downsample the LiDAR point cloud and generate uncertainty maps. Next, we grow Gaussians into incomplete regions through our focused surfel splatting technique. Finally, we extract and sample the reconstructed surfels in these regions to generate points, which integrate seamlessly with the LiDAR point cloud.}%
    \label{fig:pipeline}%
\end{figure*}
\subsection{LiDAR Artifacts}

\mainparagraph{Small geometric structures and silhouettes}
Since each light ray emitted by the sensor is not infinitesimally small, the footprint at the observed surface is proportional to the range (\textit{beam divergence}).
For example, the \textit{Velodyne VLP-16 (Lite)} sensor exhibits a beam divergence of 0.17$^\circ$~$\times$~0.09$^\circ$, resulting in a laser spot size of 40mm~$\times$~25mm at a distance of 10m~\cite{vlp16}.
If this ray, which can also be thought of as a frustum, hits a small geometric structure or an edge, the footprint of the ray becomes discontinuous.
In this case, the computed range can be inaccurate, potentially yielding an average of the observations, called \textit{mixed pixels}~\cite{tang2007comparative}. This case is shown in Fig.~\ref{fig:lidar_arti}b) and an example is given in Fig.~\ref{fig:lidar_arti}c).
This issue can be mitigated by applying a filter on the LiDAR's return type, such as prioritizing the first return or the strongest return.
However, the problem of a false measurement persists when the return is counted as a single one.
Higher channel LiDAR scanners reduce mixed pixels by lowering the signal-to-noise ratio but are significantly more costly~\cite{you2022up}.

\mainparagraph{Surface characteristics}
Dark surfaces cause weaker return signals, making them more error prone.
Since the signal-to-noise ratio is lower in this case, the range estimation becomes less reliable.
If the returned signal is too low to distinguish it from background noise, the measurement is missed completely.

Reflective or transmissive surfaces, such as mirrors or windows, generate virtual observations.
In point clouds, this often results in mirrored virtual rooms.
Whether the surface itself is detected depends on its reflectance.
This reduces measurement reliability and may result in missing sections or a high level of noise.
\new{We document these failure modes for completeness, but note that we do not specifically target error caused by dark or reflective surfaces: photometric reconstruction methods face the same challenge in these regions if they are large and without contrast, and our contribution focuses on thin structures and geometric discontinuities where RGB image cues remain informative, even if they are dark.}

\subsection{Scanner-Side LiDAR Pre-processing}\label{sec:lidar_preprocess}
\new{This pre-processing step is performed by the LiDAR manufacturer or scanner software and is typically outside the user's direct control.}
To generate reliable data for a user, the raw LiDAR data is typically pre-processed in order to remove inaccurate measurements.
Approaches to remove outliers such as mixed pixels and weak returns typically feature local neighborhood analysis~\cite{tang2007comparative}.
However, methods are vendor-specific and often proprietary, and raw LiDAR point clouds are often unobtainable.

Unfortunately, by removing points rather generously, it is accepted that some correct measurements are also lost.
Ultimately, the combination of filtering and low-confidence measurements leads to missing small and thin structures in the final point cloud.
The amount of missing points in these regions is significantly increased if their complex geometry is combined with a dark or reflective surface.
In our observation, all commonly used LiDAR datasets~\cite{Liao2022kitti, yeshwanthliu2023scannetpp,Knapitsch2017} in academic research exhibit these artifacts, with small structures and edges missing.

Thus we identify the final point cloud generally having three types of regions: \textit{well-scanned areas} (e.g.\ white, flat surfaces), \textit{gaps} (e.g.\ thin structures), and \textit{transition regions} between these two areas.
\new{Well-scanned areas pose no problem, as their measurements are reliable and dense.
Gaps are fundamentally difficult to localize: absent geometry leaves no signal in the point cloud, making a missing structure indistinguishable from intentionally empty space without external reference.
Transition regions, however, retain a characteristic low but non-zero point density, a detectable footprint produced by beam divergence and aggressive pre-processing filtering, which can be identified and exploited as candidate sites for completion.}

\section{Method}
\label{sec:method}

\new{Our aim is to complete LiDAR point clouds suffering from gaps through beam divergence or small dark surfaces. For this, we introduce a method we call \textit{SurfFill.}}
\new{We first identify transition regions through local point cloud density estimation and derive complementary uncertainty maps from monocular surface normals in the RGB images (Sec.~\ref{sec:ambiguity_heuristic}).}
Following, we pre-process and structurally downsample the LiDAR data so that \new{ambiguous regions retain high point density while well-scanned regions are thinned out}, and initialize a Gaussian surfel model with them (Sec.~\ref{sec:preprocess}).
We then grow Gaussians into missing regions by constraining 2D Gaussian Splatting optimization~\cite{huang20242d} to favourably reconstruct missing areas of the LiDAR scan (Sec.~\ref{sec:focusedopt}).
After training, we extract the splats from our total splat distribution which are most likely part of missing structures in the initial LiDAR data.
We sample multiple points from the selected splats and combine them with the LiDAR point cloud (Sec.~\ref{sec:filtering}).
An overview of our pipeline is shown in Fig.~\ref{fig:pipeline}.

\subsection{Region Ambiguity}
\label{sec:ambiguity_heuristic}
\vspace{1mm}

We identify ambiguous regions in the point cloud and RGB images \new{where beam divergence artifacts may have occurred.
Additionally, the aim is to find relevant seeding point for point growing via Gaussian Splatting.
}

\subsubsection{Ambiguity Heuristic in 3D}

Points in well-scanned regions are unlikely to get eliminated during scanning or filtering, thus they have uniformly high point densities, while regions with missing structures and empty areas have a point density of zero.
Identifying the full extent of missing areas is challenging, as it is unclear beforehand whether regions are missing or should be empty.
Transition regions, however, show a low, non-zero point density. We exploit this property to detect them in our \textit{ambiguity} heuristic.
Identifying points at the border of missing structures then enables point-growing strategies for completion.

For each point $\mathbf{x}$ of the point cloud $\mathcal{X}$, we compute an ambiguity score which we determine by the inverse density $p$ estimated using a $k$-nearest neighbor search:
\begin{equation}
    p(\mathbf{x}) = \frac{\sum^k_{i=1}\|\mathbf{x}-\mathbf{q}_i\|}{k \cdot \delta} \quad\quad \forall  \mathbf{q}_i \in \text{KNN}_k(\mathcal{X},\mathbf{x}) \ \text{and} \  \mathbf{q_i} \neq \mathbf{x} ,
\end{equation}
\new{with $\delta$ a normalization constant representing the typical inter-point distance in well-scanned regions for the given scan. In practice, $\delta$ is informed by the known LiDAR acquisition parameters, for example the target point spacing of the scanner's pre-processing pipeline (see Sec.~\ref{sec:lidar_preprocess}). It can also be estimated from point cloud statistics, though direct neighbor density estimation on real scans tends to yield slightly conservative values due to non-uniform density. We use $k=3$ for the KNN search, employing the efficient \textit{simple-knn}~\cite{kerbl20233d} framework.}
Hereby, high ambiguity represents transition areas, while low ambiguity occurs in well-scanned regions.

\new{The density-based ambiguity heuristic we introduce exploits a key property of the artifact patterns: in well-scanned regions, point clouds are dense and spatially consistent because the laser beam footprint falls entirely within a single surface. Transition regions adjacent to gaps instead exhibit characteristically low but non-zero density, as partial beam divergence and aggressive pre-processing filtering thin geometry but do not completely remove the scan near discontinuities. Crucially, the KNN-based density estimate is \textit{local}: it responds to the relative spacing of a point's $k$ nearest neighbors, not to global scanning density gradients. Moreover, such global gradients are rare in scanner-preprocessed point clouds: raw LiDAR data is commonly subsampled to a target uniform density (e.g.\ via voxel downsampling) to reduce the massive raw sample count and reach a consistent spatial resolution. The resulting near-uniform density allows the $\delta$ normalization to absorb any residual variation. While isolated occlusions or scanning irregularities can still produce false positives, our method addresses this conservatively: we identify \textit{candidate} ambiguous regions in 3D and subsequently in 2D (see next Section), however photometric reconstruction still decides specifics (see Sec.~\ref{sec:focusedopt}). Cross-sensor robustness is further supported empirically: our evaluation spans three distinct LiDAR platforms (NavVis VLX 3, FARO Focus Premium, and FARO Focus 3D X330 HDR) with substantially different beam geometries and scanning configurations, and the heuristic identifies transition regions consistently across all of them (Sec.~\ref{sec:evaluation}).}

\subsubsection{Uncertainty Maps}
\label{sect:uncertainty}
For optimization, we also want to identify ambiguous or uncertain regions in the camera images.
To this end, we establish a similar heuristic in pixel space using the differences in image-space normals.
While point densities cannot be used directly in image space, we note that density computations are commonly also used for normal estimation in point clouds through curvature.
Therefore, abrupt changes in normals in image space also can represent transition areas.

In other words, regions without differences in normals represent flat areas well scanned by the LiDAR, while large deviations represent ambiguity and thus uncertainty.
To estimate this uncertainty, we use the method of Bae et al.~\cite{bae2021estimating} as also used by Xiang et al.~\cite{xiang2024gaussianroom} to estimate surface normals.
We use the expected angular error of the normals, which represents an \textit{uncertainty} map $\mathbf{U}$, incorporated during training to focus optimization on regions of uncertainty.

\new{Large normal gradients in image space can also arise at valid, well-captured geometric edges and corners. This does not adversely affect our method: the uncertainty map is purely a guiding signal that directs photometric optimization toward regions of interest; it does not determine what geometry is ultimately accepted. If the map highlights a region that is already well-scanned, the photometric reconstruction produces correct geometry there as well. The photometric loss and subsequent opacity filtering are the true arbiters: Gaussians that grow into regions inconsistent with the image data remain nearly transparent and are discarded. False positives in the uncertainty map therefore carry no geometric consequence.}

\new{An alternative approach would be to assign RGB colors to LiDAR points by projection and detect uncertainty from color inconsistency across views. However, LiDAR point clouds frequently do not carry RGB color channels, and color-based uncertainty is sensitive to illumination differences between views. A depth-comparison approach (projecting the LiDAR cloud into camera views and finding pixels where projected depth diverges from the image content) faces the additional difficulty that missing geometry produces no projected signal, making it circular for locating precisely the absent regions. We instead derive uncertainty from monocular surface normals estimated per image: this is self-contained, requires no colored point cloud, and operates purely on image-space gradients independent of LiDAR completeness. Additionally, the Gaussian splatting optimization acts as an implicit cross-modal check: Gaussians that grow into regions inconsistent with the visual data are penalized by the photometric loss, remain nearly transparent, and are removed by the opacity filter.}

\subsection{Pre-processing}
\label{sec:preprocess}

Before reconstruction, we compute the ambiguity $p(\mathbf{x})$ per point and estimate uncertainty in the camera views, as described in Sec.~\ref{sec:ambiguity_heuristic}.
Next, we downsample our input point cloud non-uniformly.
High ambiguity points with $p(\mathbf{x}) > \tau$ are kept in full, while the rest is downsampled randomly.
\new{We set $\tau = 0.04$, determined via an ablation study on the threshold sensitivity (see \ref{supp:hyperstudy}).}
As depicted in Fig.~\ref{fig:pipeline}, this results in a point cloud in which ambiguous transition regions retain high density while well-scanned regions are thinned out.

\new{Although thin structures themselves may already be partially absent in the LiDAR scan, their surrounding transition-region points (whose high ambiguity scores ensure they survive downsampling) guide subsequent Gaussian growing into the correct spatial locations. Well-scanned regions, on the other hand, have low ambiguity scores and are initialized with precise LiDAR measurements. Since our focused optimization naturally concentrates growth on high-ambiguity regions and leaves well-scanned Gaussians under a stricter pruning threshold (see Sec.~\ref{sec:focusedopt}), aggressively downsampling well-scanned areas does not hurt reconstruction quality there; it only increases the capacity available for point growing in incomplete regions, as demonstrated in Fig.~\ref{img:uniform_sampling}. Importantly, any residual noise in the final output is therefore confined to transition zones adjacent to newly reconstructed structures; well-scanned flat regions are shielded by the distance-based filter (Sec.~\ref{sec:filtering}), which discards any Gaussian closer than $t_{min}=0.01$\,m to the original scan.}

This results in a strong reduction of the point cloud, but in some scenes the resulting point clouds still exceed GPU memory limitations (which for us is around 2 million points).
In such cases, we use a chunking approach as described in Sec.~\ref{sec:largescale}.

We initialize the Gaussian surfel model and set both initial scales $s=s_u=s_v$ to be extra small with $s_{\mathbf{x}} = \sqrt{p(x)} \cdot 2^{-1/2}$, stimulating finer reconstruction and earlier point growth when densifying these regions.

\subsection{Focused Gaussian Surfel Optimization}
\label{sec:focusedopt}

We focus our Gaussian surfel model on incomplete areas.
As such, we introduce adjustments to densification and pruning and three additional constraints to optimization to help guide reconstruction for fine details in missing LiDAR areas.

\begin{figure}
    \centering%
    \includegraphics[width=1.0\linewidth]{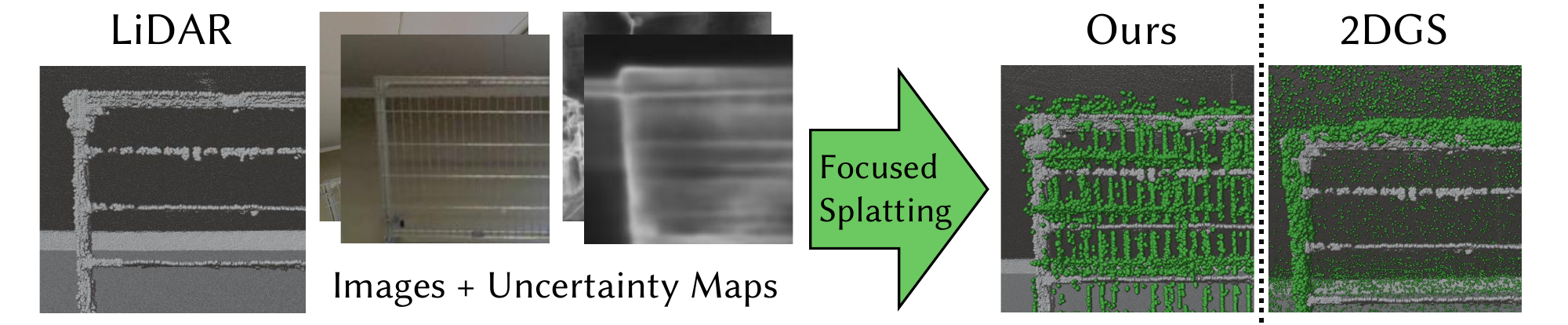}%
    \caption{Focused Gaussian surfel optimization. Our focused surfel splatting method reconstructs finer details than baseline 2DGS~\cite{huang20242d}.}%
    \label{fig:gaussian_optim}%
\end{figure}

This adjustment is necessary because standard reconstruction methods often miss precise edges and details, focusing instead on reconstructing broad flat surfaces while ignoring fine structures (refer to Fig.~\ref{fig:gaussian_optim} (right, 2DGS)). We assume high precision from the LiDAR input and need the optimization to \textit{focus} on finer details.

Additionally, we augment Gaussians with their point ambiguity, which is repeatedly updated during training and used during optimization as well as filtering.

\mainparagraph{Densification and pruning}
We require point growth to handle both small details and large missing structures, such as entire table legs (see Fig.~\ref{fig:teaser}).
Therefore, we adjust the densification strategy from the Adaptive Density Control (ADC) module~\cite{kerbl20233d}.

First, we adopt the noise term from Kheradmand et al.~\cite{kheradmand20243d} for Gaussian surfel splatting.
In each iteration, we add a small jitter in the tangent vector plane to the surfel's positions (see \ref{supp:noise_details} for details).
However, we rely on ADC as the base model, as this proved more robust for our task (Sec.~\ref{sec:ablations}).

Furthermore, we employ the maximum instead of the mean screen-space gradient as the heuristic for splitting and cloning~\cite{kerbl2024hierarchical}.
This improves results for the sparse RGB datasets often captured with LiDAR point clouds.
We also pass the ambiguity score during ADC to the child and note the Gaussian as having been newly created with a ``was-densified'' flag. This lets us identify novel surfels during filtering.

For pruning, points with low ambiguity are assumed to be accurate, as well-scanned areas have low densities after downsampling.
They have precise initializations from the LiDAR data and should not be removed during optimization.
Thus, they are only pruned if their opacity becomes 100 times lower than the standard pruning threshold~\cite{kerbl20233d}.

\mainparagraph{Training view sampling}
In 3DGS~\cite{kerbl20233d}, training views are uniformly sampled. This is suboptimal for us, as views of e.g.\ flat surfaces are not necessary for LiDAR completion.
By counting the number of projected high-ambiguity points $n_{a}$ in each image, we adjust training view sampling probability to
    $P_{image} = n_{a}/{|\mathcal{N}_\text{v}|}$,
where $|\mathcal{N}_\text{v}|$ is the number of Gaussians in each camera's frustum.

\begin{figure*}
    \centering%
    \includegraphics[width=0.7\linewidth]{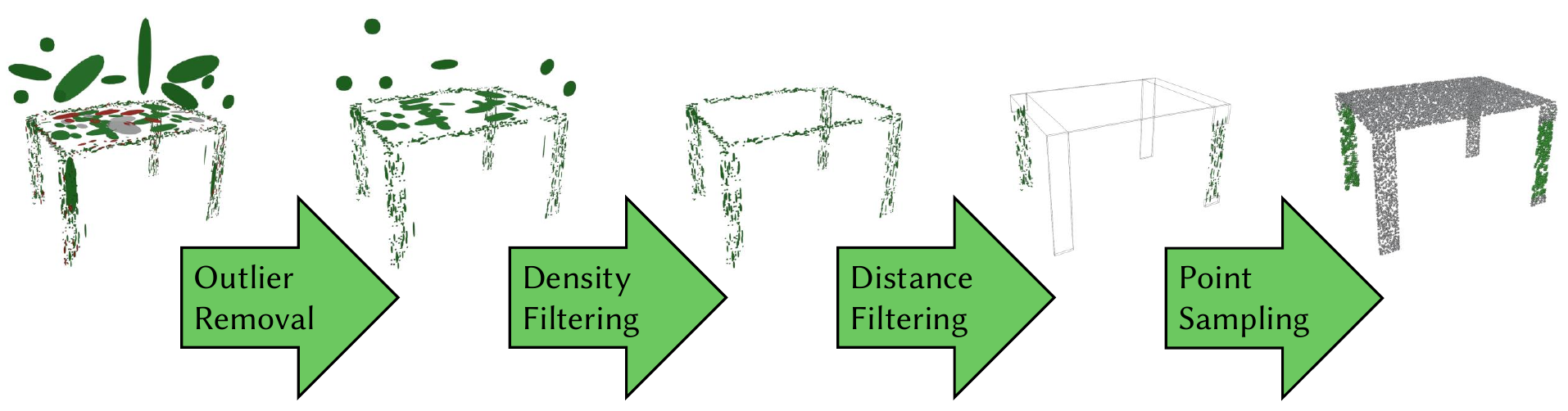}%
    \vspace{-2mm}
    \caption{Our filtering strategy. We first remove outliers, nearly transparent, large splats or non-densified splats. Next, high-ambiguity Gaussians surfels are removed and the model is filtered based on their distance to the input point cloud. Finally, multiple points are sampled per surfel.}%
    \label{fig:filtering}%
\end{figure*}

\mainparagraph{Losses and regularizers}
Gaussian-based geometry reconstruction methods~\cite{huang20242d,dai2024high} aim
to reconstruct entire scenes for which a balanced set of losses and regularizers are used.
In contrast, our method aims to accurately reconstruct LiDAR artifacts, for which we adjust the optimization target.

We focus on finely reconstructing edges~\cite{Gong2024EGGSEG} with a gradient loss $\mathcal{L}_G$ between rendered image $\mathbf{I}$ and ground truth $\mathbf{I}_{GT}$:
$\mathcal{L}_g = \|\nabla \mathbf{I} - \nabla \mathbf{I}_{GT} \|$.
Compared to an explicit Canny edge detector and loss, this proved more robust and faster to compute.
Furthermore, we introduce a scale regularization $\mathcal{R}_s$ for all Gaussians $\mathcal{N}$ to ensure small-detailed point reconstruction with
\begin{equation}
    \mathcal{R}_s = \frac{1}{2 |\mathcal{N}|}  \sum\nolimits^{|\mathcal{N}|}_{i=1} (s_{u,i} + s_{v,i}) \ .
\end{equation}
This also causes better primitive distributions for accurate filtering.

Additionally, the computed 2D uncertainty maps $\mathbf{U}$ are used to further focus the optimization on incomplete areas.
We create binary masks $\mathbf{M}$ (and the inverse $\overline{\mathbf{M}}$) by applying a threshold of \new{$0.2$} (see Sec.~\ref{sec:ablations}), such that $\mathbf{M}$ includes regions of interest for completion and $\overline{\mathbf{M}}$ accurately scanned ones, yielding our objective function:
\begin{equation}
    \mathcal{L} = (0.5 + 0.5 \cdot \mathbf{M}) (\mathcal{L}_{c}+ \gamma\mathcal{L}_g)+\rho \mathcal{R}_s + \overline{\mathbf{M}}( \alpha\mathcal{L}_d  + \beta \mathcal{L}_n).
\end{equation}
As in 3DGS~\cite{kerbl20233d}, $\mathcal{L}_{c}$ is the weighted combination of color losses $\mathcal{L}_1$ and D-SSIM; however, we increase D-SSIM weighting with $\lambda=0.4$ to penalize inaccurate edges.
Both $\mathcal{L}_d$ and $\mathcal{L}_n$ are the depth distortion and normal consistency losses from Huang et al.~\cite{huang20242d}.
However, they are masked to already accurate regions as they encourage smooth surfaces and work best for flat areas.
Otherwise, fine structures would be lost.
As seen in Fig.~\ref{fig:gaussian_optim}, the reconstruction thus provides a well-rounded Gaussian model for point completion via filtering and sampling.

\subsection{Gaussian Surfel Filtering and Sampling}\label{sec:filtering}

After training, the Gaussian surfel model is filtered and sampled for point completion.
See Fig.~\ref{fig:filtering} for a visualization of our filtering steps.

\mainparagraph{Filtering}
We first remove Gaussians with at least one scale larger than a threshold based on the scan, e.g.\ $10\times$, thus 5\,cm for 5\,mm point cloud resolutions.
These are outliers, floaters, or represent large, well-scanned areas, and thus irrelevant for completion.
We also discard low-opacity Gaussians, as they do not accurately represent a surface~\cite{yu2024gaussian}.

Next, original Gaussians via the ``was-densified'' flag as well as Gaussians with remaining high ambiguity are removed, as they reconstruct unimportant regions.

The last filter step involves comparing each Gaussian's position to the input LiDAR point cloud.
We keep clusters of points away from scanned structures, thus we select Gaussians which fulfill:
\begin{equation}%
  \begin{aligned}
    t_{min} <  \sum\nolimits_{i=1}^k \| \mathbf{p} - \mathbf{q}_i \| < t_{max}, \\
    \quad \forall  \mathbf{q} \in \text{KNN}_k(\mathcal{N},\mathbf{p}) \ \text{and} \  \mathbf{q} \neq \mathbf{p}
      \end{aligned}
\end{equation}%
\new{with $k=5$,} $t_{min}=0.01$\,m and $t_{max}=3$\,m used for all test scenes.
If the position is closer than $t_{min}$, we have duplicated information and remove the Gaussian, as we assume the LiDAR scan to be more accurate.
If it is further away than $t_{max}$, the Gaussian is an outlier.
\new{
This distance-based filter is also the primary mechanism preventing the introduction of noise into previously clean LiDAR regions: only Gaussians that are spatially away from the original scan, and thus represent genuinely new geometry, are retained for sampling.}

\mainparagraph{Sampling}

The filtered Gaussian model is sampled for points.
First, in addition to the surfel's center point, we sample the Gaussian distribution of each surfel.
Second, points are sampled between two Gaussians to bridge large gaps in thin structures with
\begin{equation}
   p' = (1-a) \mathbf{p} + a \mathbf{q}_i \quad \mathbf{q}_i \in \text{KNN}_k(\mathcal{N},\mathbf{p}) \ \text{and} \  \mathbf{q} \neq \mathbf{p}.
\end{equation}
Here, $i$ is a randomized neighbor index and $a$ is a randomized distance.
This is necessary if few Gaussians represent a structure. It works as we kept clusters during filtering.
Both of these strategies allow selecting a desired amount of points per Gaussian, and thus integrate well with the point density of the scan.

\new{\mainparagraph{Point normals}
Input LiDAR points carry per-point normals, often estimated via KNN in the scanner pre-processing pipeline. For newly sampled points, our implementation does not output normals. Two natural extensions exist: normals could be inherited from the parent Gaussian surfel via $\mathbf{n} = \mathbf{t}_u \times \mathbf{t}_v$ (where $\mathbf{t}_u$, $\mathbf{t}_v$ are already optimized for geometric consistency during training), or estimated by applying the same KNN approach on the merged point cloud. Either would provide per-point normals for completed points suitable for downstream applications such as meshing or rendering.}

\section{Large-Scale Completion}
\label{sec:largescale}

\begin{figure}
    \centering%
    \includegraphics[width=1\linewidth]{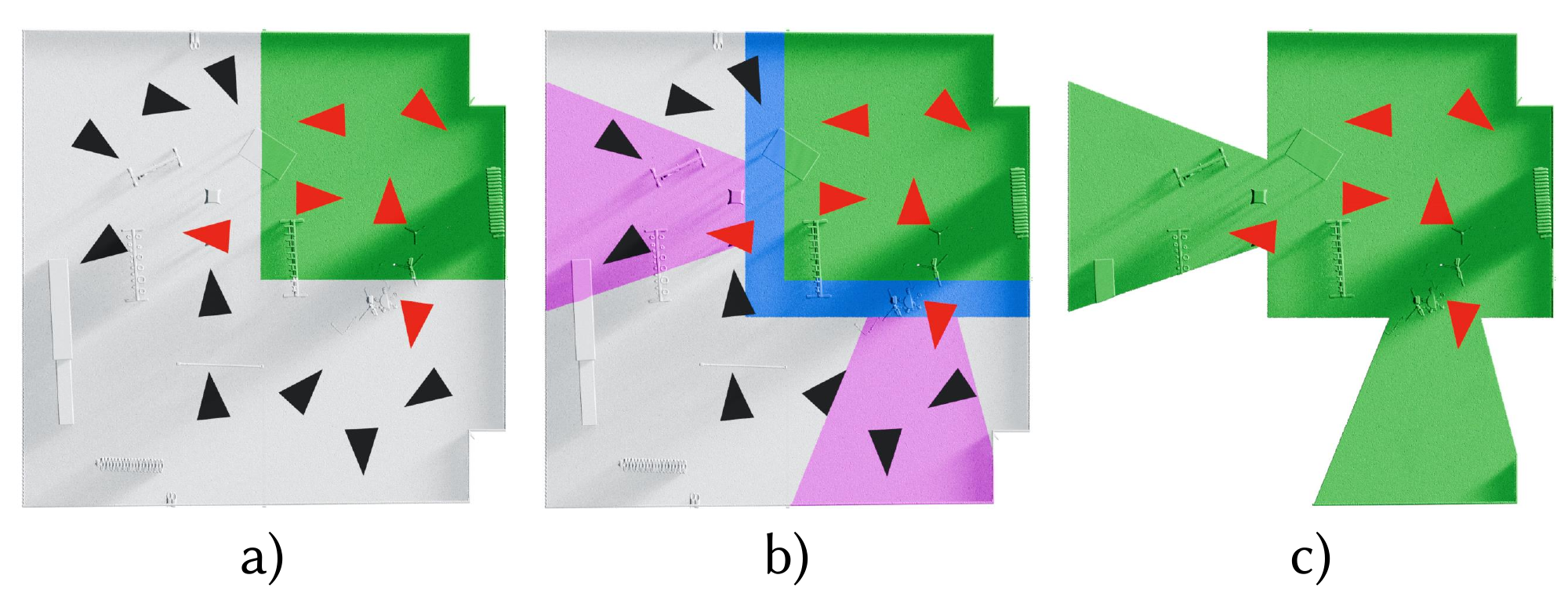}%
    \vspace{-2mm}
    \caption{The handling of large scenes shown with an example room. a) A subset of cameras (red) for a chunk (green) are selected. All cameras within the chunk's bounding box and those seeing the chunk at a user defined distance are chosen. b) Points for the chunk are selected with the point set within the bounding box (green) extended by 20\% (blue) and by all points inside the cameras' frusta (pink). c) The resulting chunk's dataset.}%
    \label{fig:large_scale_scheme}%
    \vspace{-2mm}
\end{figure}

LiDAR captures contain more points than common object-centric Gaussian surface reconstruction test scenes.
These scans can include tens of millions of individual points and datasets can consist of multiple rooms or buildings.
For accurate point cloud completion, we apply a divide-and-conquer technique to our method.
Otherwise, reconstruction with Gaussians is limited to optimizing a few million primitives at most due to GPU memory constraints.
We follow the proposed chunking approach described in Kerbl et al.~\cite{kerbl2024hierarchical} and Lin et al.~\cite{lin2024vastgaussian}.
The method divides the dataset into axis-aligned square boxes and then selects a set of training image cameras as well as a subset of the point cloud for each chunk.
This allows independent processing of each chunk.
A visualization of the process can be seen in Fig.~\ref{fig:large_scale_scheme} and details can be found in the mentioned prior works.

After chunking, each part is trained on one GPU, allowing parallel computation of all chunks when using multiple GPUs.
After reconstruction, Gaussians outside the original square bounding box are discarded before we commence our normal filtering pipeline.

For details on chunking and overlapping, see \ref{supp:large_scale} and \ref{supp:chunking_artifacts}.

%% file: 04-eval.tex
\newcommand{\evalparagraph}[1]{\paragraph*{#1}}

\section{Evaluation}
\label{sec:evaluation}

\begin{figure*}[]
	\centering
	\includegraphics[width=0.95\linewidth]{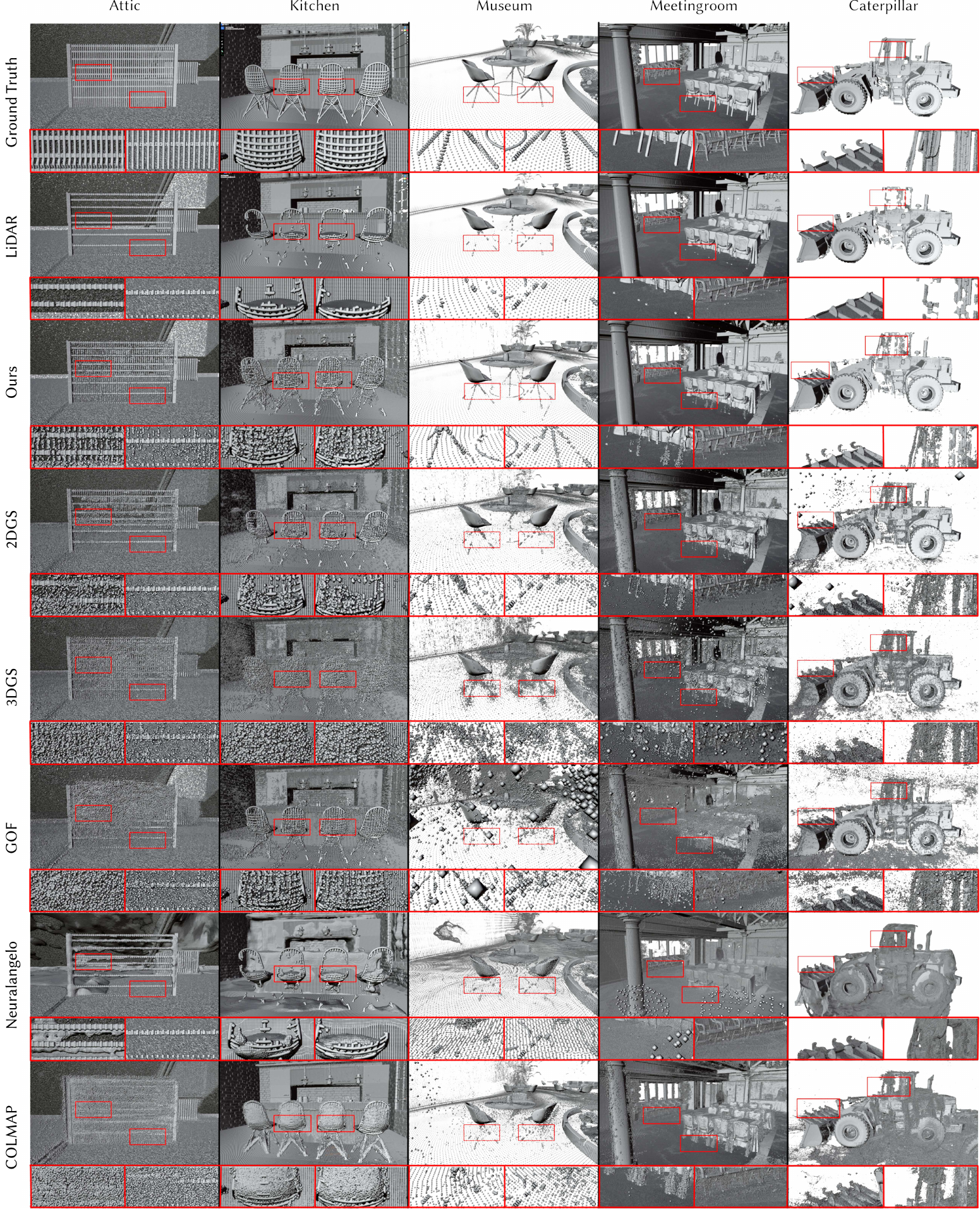}
	\caption{Resulting point clouds for our test scenes trained using all evaluated methods. Our method is best able to reconstruct the small structures missed.}
	\label{img:comparison_grid}
\end{figure*}
\begin{figure*}[]
	\centering
	\includegraphics[width=0.95\linewidth]{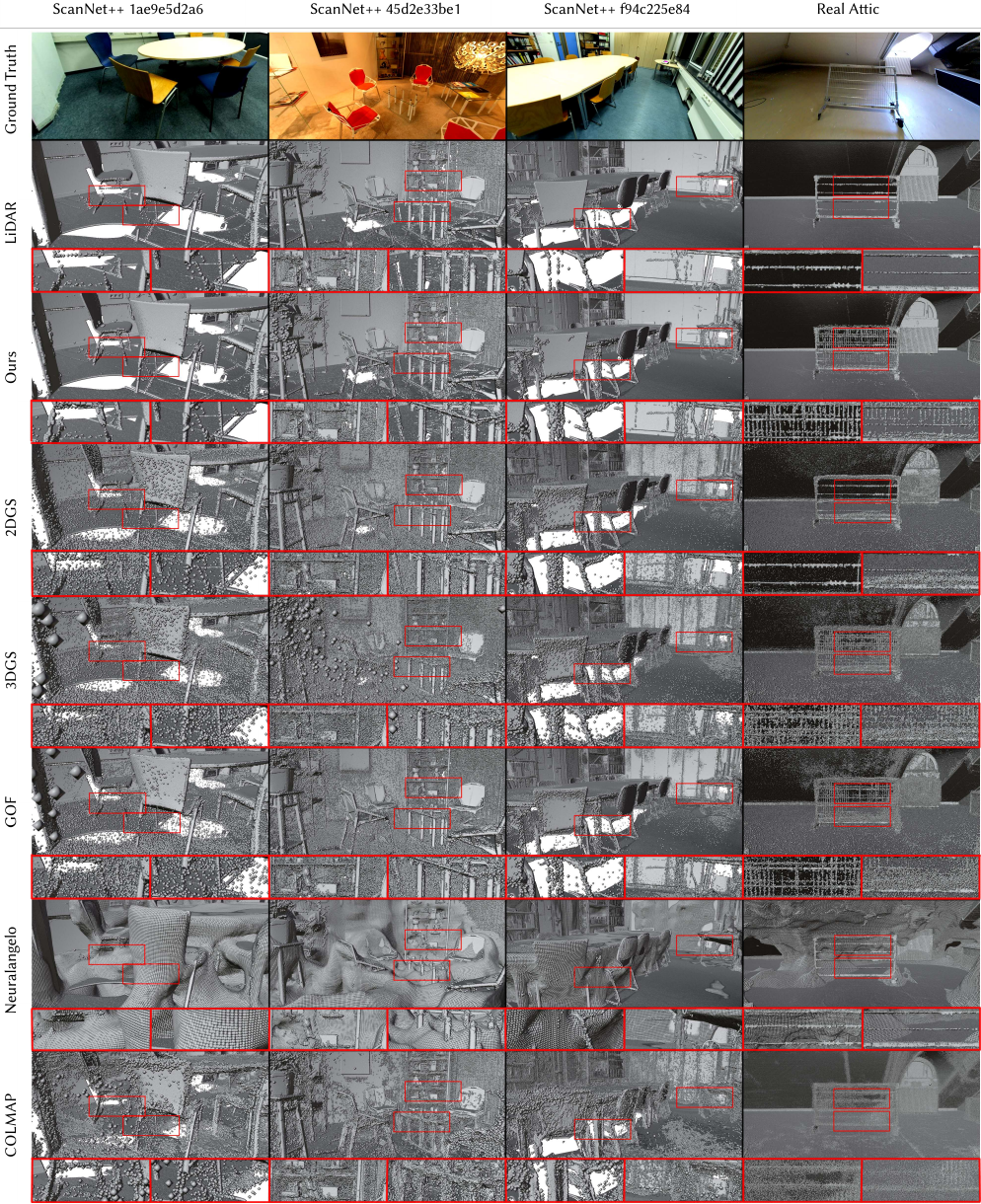}
	\caption{Resulting point clouds for real-world datasets trained with the evaluated methods: The scenes from ScanNet++ are scanned with a FARO Focus Premium LiDAR, the \textsc{real attic} scene is generated with the NavVis VLX 3. Our method completes the small structures missed by the LiDAR sensors best.}
	\label{img:comparison_grid_realworld}
\end{figure*}

Evaluating the geometrical correctness of LiDAR completion is challenging due to the lack of real-world ground-truth datasets.
To address this, we use two types of datasets.
\new{Our evaluation is structured to test each contribution directly: Sec.~\ref{sec:ablations} ablates the individual components of the SurfFill pipeline; the quantitative and qualitative comparison that follows evaluates performance against baselines across the full synthetic and real-world test suite; and the Further Experiments subsection probes specific contributions including the initialization strategy, large-scale chunking, pose robustness, and comparison with direct shape completion methods.}

\evalparagraph{Test scenes}
We evaluate on three synthetic scenes \textsc{attic}, \textsc{kitchen} and \textsc{museum} with a small pose noise from a Gaussian distribution of spread $0.01^{\circ}$ added, reflecting industrial scanning accuracy~\cite{lixell2}. 
\textsc{attic} is modeled after a real-world scan.
Points are removed based on typical scanning artifacts (see Sec.~\ref{sec:lidar_review}, visualizations in \ref{supp:test_scene_details}).
Furthermore, we evaluate on \textsc{meetingroom} and \textsc{caterpillar} from the Tanks\&Temples dataset \cite{Knapitsch2017}, which are the most complete available scans. 
We use the same point removal scheme. However, we do not add pose noise and compare against the complete scan.

\evalparagraph{Real-world scenes}
For real-world evaluation, we use three of our own captured scans \textsc{real attic}, \textsc{office} and \textsc{bridge} %
, three scenes from the ScanNet++ dataset~\cite{yeshwanthliu2023scannetpp} and two additional scenes from Tanks\&Temples.
An overview of the characteristics and LiDAR hardware of the scenes can be found in Tab~\ref{tab:scenes}.
For captured images, both our own and ScanNet++ use fisheye images, which we undistort to 90$^\circ$ pinhole.

\evalparagraph{Quantitative performance}
We assess performance using the Chamfer Distance as well as the F1-score metric~\cite{Knapitsch2017}, combining \textit{precision} (reconstruction accuracy) and \textit{recall} (coverage) using a 5mm threshold. %
Our goal is to recover details missed in the scans, which often account only for small differences in the F1-score.
Nonetheless, evaluating the entire scene is crucial to ensure reconstruction quality does not degrade in well-scanned areas. 

\subsection{Ablations}
\label{sec:ablations}
\begin{table}
\centering
\caption{\label{tab:ablation_study}%
F1-scores and visualizations for test scene ablations: 'w/o Maximum Gradients' indicates the use of mean gradients, while 'w/o Calculated Probabilities' denotes the use of a uniform probability distribution.}
\small
\renewcommand{\arraystretch}{1} %
\setlength{\tabcolsep}{2pt} %
\begin{tabular}{l|ccc|c}
\textbf{} & \textsc{Attic} & \textsc{Kitchen} & \textsc{Caterpillar} & Avg. \\\hline
\text{Full Model} & 0.9509 & 0.8419 & 0.9223 & 0.9049 \\\hline
\text{w/o Filtering} & 0.9431 & 0.8224 & 0.8830 & 0.8828 \\
\text{w/o Maximum Gradients} & 0.9489 & 0.8347 & 0.9217 & 0.9018 \\
\text{w/o Positional Noise} & 0.9499 & 0.8378 & 0.9218 & 0.9032 \\
\text{w/o Scale Regularization} & 0.9500 & 0.7836 & 0.9217 & 0.8851 \\
\text{w/o Uncertainty Maps} & 0.9503 & 0.7878 & 0.9199 & 0.8860 \\
\text{w/o Calculated Probabilities} & 0.9503 & 0.8430 & 0.9222 & 0.9052 \\
\text{w/o Edge Loss} & 0.9500 & 0.8409 & 0.9224 & 0.9044 \\
\text{w/o Stricter Pruning} & 0.9505 & 0.8410 & 0.9222 & 0.9046 \\
\end{tabular}
\includegraphics[width=1\linewidth]{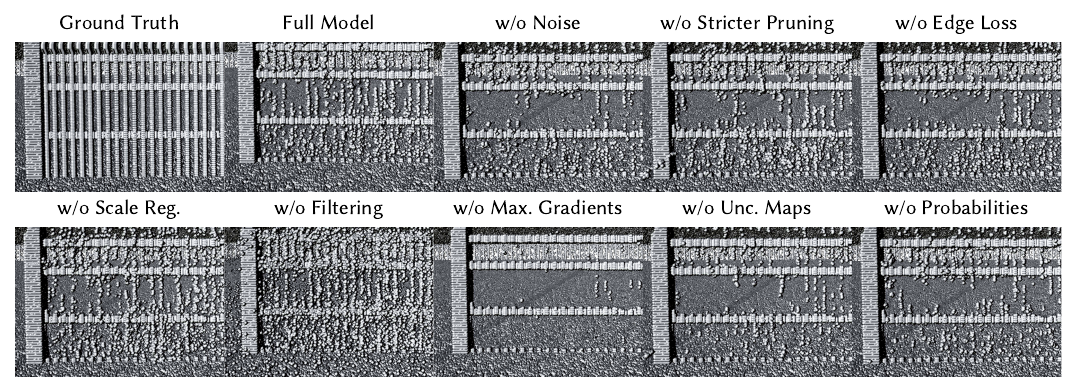}
\end{table}

We conducted ablation studies to assess the impact of individual components using the F1-score across three test scenes: \textsc{Attic}, \textsc{Kitchen}, and \textsc{Caterpillar}.
For quantitative and qualitative results, see Tab.~\ref{tab:ablation_study}.
Our filtering method proves impactful, increasing the F1-score on average by 2.5\%.
Using maximum gradient densification~\cite{kerbl2024hierarchical} also leads to improvements in all scenes.
Adding positional noise for exploration further increases the F1-score by up to 1.6\% and improves the reconstruction of fine details in the \textsc{Attic}. 
Using the full densification scheme of Kheradmand et al.~\cite{kheradmand20243d}, however, leads to a decrease in geometric quality. For example, the F1-score on the \textsc{Attic} drops from 0.9509 to 0.9460.
Scale regularization, while having minor effects on \textsc{Attic} and \textsc{Caterpillar}, prevents excessive splat generation to reconstruct the brick wall texture, stabilizing \textsc{Kitchen} and improving F1-scores by 7.3\%.
Similarly, uncertainty maps disincentivize excessive Gaussian growing, reducing errors in textured regions.
Lastly, our probability distribution and edge loss primarily target small structures.
Although their impact on scores is minimal or slightly negative on \textsc{Kitchen}, they are essential to reconstruct finer details, as seen in the qualitative comparisons included in Tab.~\ref{tab:ablation_study}.
Together, all parts improve the point cloud's quality, especially in challenging regions.

\new{The ablation also highlights the role of the filtering step ($+2.5\%$ F1 on average) as the primary mechanism keeping noise out of well-scanned regions: without filtering, many Gaussians that grew into already-accurate LiDAR areas are included, introducing spurious points. The distance-based filter (Sec.~\ref{sec:filtering}) removes Gaussians that are too close to the original scan, directly limiting noise in flat regions.}

\begin{table*}[]
\centering%
\caption{\label{tab:comparison}Chamfer Distance (m), F1-Score and runtime results of our test scenes enhanced using different LiDAR completion methods: Our method achieves the highest performance, followed by using 2DGS and 3DGS respectively. Our changes to baseline 2DGS do not lead to significantly slower computation times. The best score for each test scene is highlighted in green, while the second-best is marked in yellow. Time is measured on an Nvidia A40 GPU.}%
\renewcommand{\arraystretch}{1.0} %
\setlength{\tabcolsep}{2pt} %
\makebox[0pt][c]{\parbox{1.0\textwidth}{%
\vspace{5mm}
\begin{minipage}[t][][b]{0.75\linewidth}%
\small

\centering
\renewcommand{\arraystretch}{0.8} %
\resizebox{\textwidth}{!}{\begin{tabular}{l|c|ccccc|c|c|r}
Completion & Large& \multicolumn{6}{c|}{Chamfer Distance (m) $\downarrow$} & F1-score $\uparrow$\\
\text{Method} &  Scale & \textsc{  Attic  } & \textsc{  Kitchen  } & \textsc{ Museum } & \textsc{Meetingroom} & \textsc{Caterpillar} & \text{ Mean }  & \text{ Mean } & \text{Time} $\downarrow$\\\hline
\multicolumn{10}{l}{\textit{\new{Image-only reference baselines (no LiDAR input)}}} \\\hline
\text{Neuralangelo} & $\times$ & 0.1081 & 0.0893 & 0.2795 & 1.2270 & 0.4249 & 0.4258 & 0.7765  &$\sim$24 h\\
\text{COLMAP} & $\checkmark$ & 0.0041 & 0.0148 & 0.0328 & 0.0168 & 4.8968 & 0.9930 & 0.8141  &3.7 h \\\hline
\multicolumn{10}{l}{\textit{\new{LiDAR-assisted methods}}} \\\hline
\text{3DGS} & $\times$ & 0.0034 & \cellcolor{yellow}0.0104 & 0.0145 & 0.0026 & 0.8357 & 0.1733 & \cellcolor{yellow}0.8795 &  \cellcolor{green!50}21.6 m\\
\text{2DGS} & $\times$ & \cellcolor{yellow}0.0025 & 0.0416 & \cellcolor{yellow}0.0109 & \cellcolor{yellow}0.0017 & \cellcolor{yellow}0.2526 & \cellcolor{yellow}0.0618 & 0.8781 & \cellcolor{yellow}23.7 m \\
\text{GOF} & $\times$ & 0.0039 & 0.0129 & 0.0713 & 0.0141 & 0.5199 & 0.1244 & 0.8731 &124.1 m\\
\text{Ours} & $\checkmark$ & \cellcolor{green!50}0.0019 & \cellcolor{green!50}0.0087 & \cellcolor{green!50}0.0079 & \cellcolor{green!50}0.0016 & \cellcolor{green!50}0.0071 & \cellcolor{green!50}0.0055 & \cellcolor{green!50}0.9176  & 24.9 m \\
\end{tabular}}
\end{minipage}%
\hfill%
\begin{minipage}[t][][b]{0.24\linewidth}\centering%
\vspace{-2mm}
\includegraphics[width=1\linewidth]{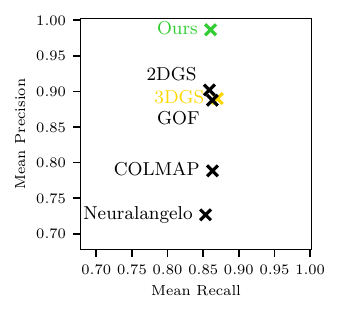}%
\end{minipage}%
\hfill%
}}%
\vspace{-2mm}
\end{table*}

\subsection{Quantitative \& Qualitative Evaluation}

We evaluate our method against a photogrammetry baseline using \textit{COLMAP}~\cite{schoenberger2016mvs} and the NeRF-based \textit{Neuralangelo}~\cite{li2023neuralangelo}.
For Gaussian approaches, we compare with \textit{3DGS}~\cite{kerbl20233d}, \textit{2DGS}~\cite{huang20242d}, and \textit{GOF}~\cite{yu2024gaussian}. We apply a uniform downsampling scheme to the input point clouds.
For filtering, we use the same opacity threshold as within our method.

\new{We include COLMAP and Neuralangelo as reference baselines representing purely image-based reconstruction, without any LiDAR input. They are not direct competitors, as our method uses LiDAR as the primary data source and complements it with photometric reconstruction, but provide important context for understanding how much purely visual methods can contribute to geometry recovery. Deviations between these baselines and the LiDAR input are therefore expected. Importantly, the precision--recall figure in Tab.~\ref{tab:comparison} reveals that COLMAP achieves a mean \textit{Recall} of 0.863, almost identical to ours (0.860): it finds a comparable number of the missing points. However, its mean \textit{Precision} is only 0.789 versus 0.986 for our method, confirming that COLMAP recovers similarly many points but at far lower spatial accuracy. This distinction is masked by the global F1-score and explains why COLMAP appears competitive on recall while underperforming on overall geometry quality. Our method's primary gain over COLMAP is therefore in precision; recall is comparable, reflecting that our photometric reconstruction prioritizes accurate surface localization over coverage in ambiguous regions.}

\new{Two recent methods that combine LiDAR with Gaussian Splatting warrant specific discussion: Li-GS~\cite{jiang2024li} and Tclc-GS~\cite{zhao2024tclc}. Li-GS uses LiDAR depth as a regularizer for surface reconstruction but explicitly restricts Gaussian primitives to the original LiDAR surface and does not fill geometric gaps, which is the core task addressed by our method. Tclc-GS targets outdoor autonomous-driving scenarios with sparse, sweep-based LiDAR patterns, a fundamentally different setup from the room- and building-scale indoor completion we address. Furthermore, neither method provides a public implementation at the time of submission, making reproducible quantitative comparison infeasible.}

\evalparagraph{Test scenes}
Results can be seen in Tab.~\ref{tab:comparison} and Fig.~\ref{img:comparison_grid}.
They show that our method excels in reconstructing fine geometry with minimal noise, evident in the \textsc{Attic} fence bars and chair legs in \textsc{Museum} and \textsc{Meetingroom}. 
However, some noise remains, particularly around the legs of the \textsc{Kitchen} chairs and \textsc{Caterpillar} tires.

Other methods struggle to improve LiDAR point clouds.
3DGS clusters Gaussians near fine structures, while 
GOF performs inconsistently. GOF excels in some areas, such as the \textsc{Kitchen} chairs but fails in others, e.g. the \textsc{Attic} fence or \textsc{Museum}.
2DGS generates less noise, but sacrifices fine structures for overall geometric consistency, as seen with the \textsc{Attic} fence bars. 
The non-Gaussian methods perform poorly. 
Neuralangelo rarely reconstructs gaps, while COLMAP shows potential, e.g. with the \textsc{Kitchen} chairs. However, it lacks fidelity, clustering points without restoring details.

\new{The scale of the quantitative improvement is notable: our method achieves a mean Chamfer Distance of 0.0055\,m, an eleven-fold improvement over the next-best Gaussian approach (2DGS at 0.0618\,m, Tab.~\ref{tab:comparison}). This gap is not attributable to any single component in isolation: the initialization experiment (Fig.~\ref{img:uniform_sampling}) shows that supplying 2DGS with our ambiguity-guided preprocessing still yields notably worse results than the full SurfFill pipeline, confirming that focused optimization and filtering each contribute independently to the result.}

\new{To shed further light on where the improvement comes from, we note that our overall gains are concentrated in regions of high geometric complexity: the \textsc{Attic} fence (precision-critical thin structures), chair legs in \textsc{Museum} and \textsc{Meetingroom}, and hooks and cabin structure in \textsc{Caterpillar}.}

\evalparagraph{Real-world scenes}
For qualitative results of our real-world scenes, see Fig.~\ref{img:comparison_grid_realworld}.
Our method introduces minimal noise to the datasets, while reconstructing fine structures. 
For instance, in the challenging \textsc{Real Attic} scene, our method successfully reconstructs most of the fence bars, outperforming other approaches that fail to do so or generate significant noise. 
In the three ScanNet++ scenes, our method excels at reconstructing small, missing elements, like the chair legs. 
Other methods introduce substantial outliers in these scenes, whereas our approach maintains a clean reconstruction.
For comparisons, please also see the supplemental video (\url{https://youtu.be/OS3q5OWT-sg}).
Further results for \textsc{Bridge} and two Tanks\&Temples scenes are in \ref{supp:tnt2} and \ref{supp:bridge}.

\subsection{Further Experiments}
\new{The following experiments each probe a specific aspect of the SurfFill pipeline: how the ambiguity-guided initialization compares to uniform downsampling (Initialization); how many of the synthetically removed points are actually recovered (Missing Points); why scene-scale LiDAR completion requires a different approach than object-centric deep completion (Shape Completion Methods); and the practical limits of the system at scale and under degraded inputs (additional experiments).}

\begin{figure}[]
	\centering
	\includegraphics[width=0.9\linewidth]{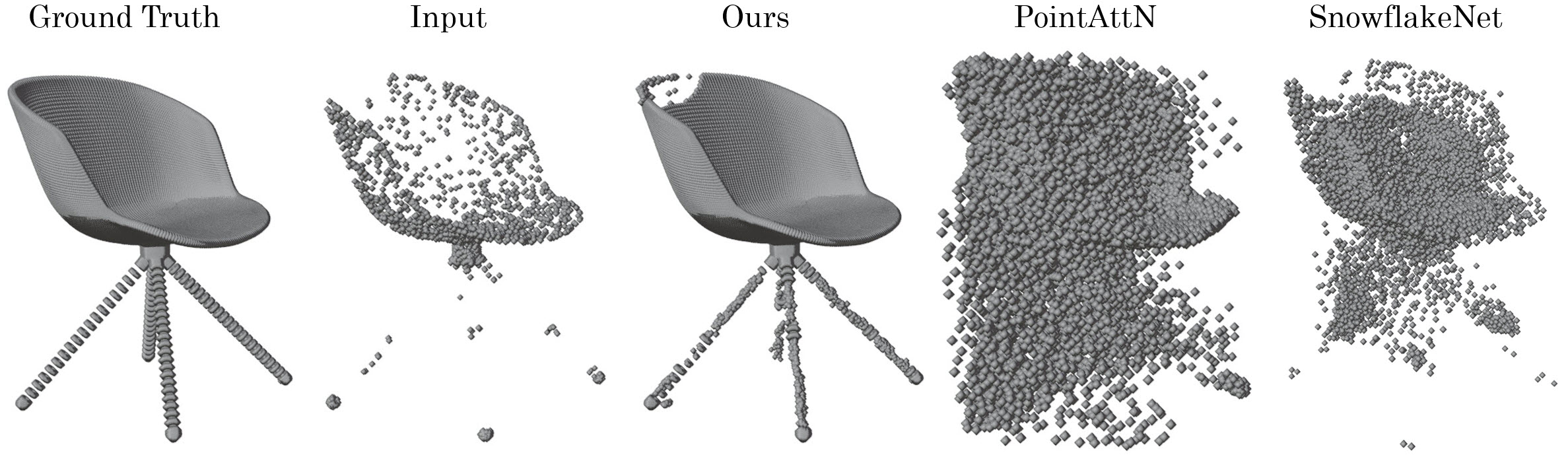}
	\caption{Comparison between our method and the direct shape completion techniques SnowflakeNet~\cite{xiang2021snowflakenet} and PointAttN~\cite{pointattn} for a simple chair example.}
	\label{img:pointattnchair}
\end{figure}
\evalparagraph{Comparison with Shape Completion Methods}

We compare our method against two leading deep learning-based completion methods: \textit{PointAttN}~\cite{pointattn} and \textit{SnowflakeNet}~\cite{xiang2021snowflakenet}. These methods were chosen due to their strong performance on the Completion3D~\cite{completion3d} and PCN datasets~\cite{pcndataset}, which are standard benchmarks for direct point cloud completion methods. However, when applied to partial LiDAR-derived models from our synthetic museum scene (chairs and tables) they generalize poorly, despite being trained on similar object categories. As shown in Figure~\ref{img:pointattnchair}, our method significantly outperforms both baselines, achieving an F1-score of 0.8918 compared to just 0.1054 for SnowflakeNet. These results highlight our method’s ability to handle complex real-world scenes. Further experimental details and quantitative results can be found in \ref{supp:direct_pc_comp}.

\begin{figure}[]
	\centering%
	\includegraphics[width=1\linewidth]{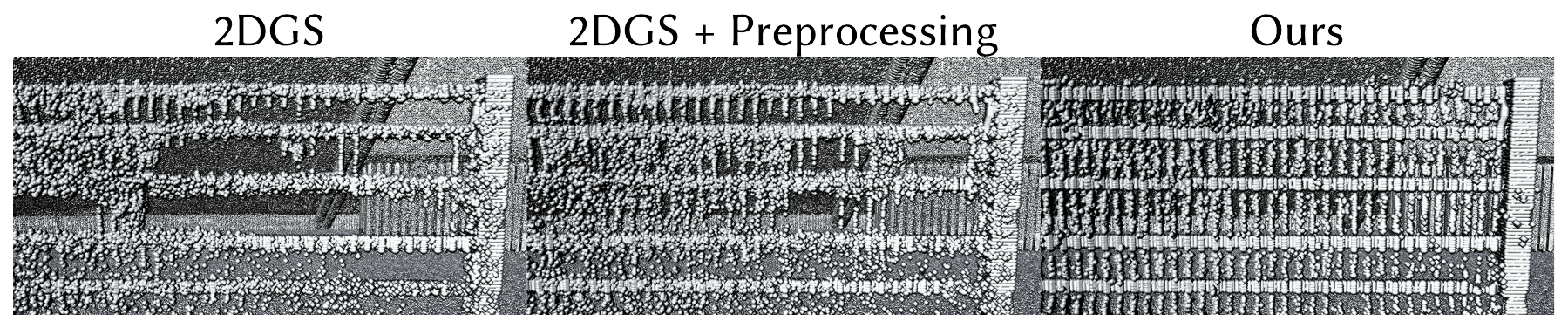}%
	\caption{Resulting point clouds for \textsc{Attic} using our method, 2DGS initialized with a uniformly downsampled LiDAR point cloud and 2DGS initialized with a LiDAR point cloud downsampled with the ambiguity heuristic.}%
	\label{img:uniform_sampling}%
\end{figure}

\evalparagraph{Initialization}
\new{This experiment isolates the contribution of the ambiguity-guided preprocessing from the focused optimization, by testing what happens when a standard method receives our improved initialization instead of a naive uniform downsampling.}
We train 2DGS with two different initializations:
First, we run 2DGS with a naively uniformly downsampled point cloud. 
We find that this leads to significantly less reconstructed beams in the \textsc{Attic} scene than if we provide 2DGS with the LiDAR point cloud structurally downsampled with our preprocessing using our ambiguity heuristic (Secs.~\ref{sec:ambiguity_heuristic} and \ref{sec:preprocess}). 
This is shown in Fig.~\ref{img:uniform_sampling}. 
\new{Our initialization increases quality drastically.}
However, even with this better initialization, 2DGS still performs notably worse than our focused 2DGS approach\new{, showcasing the effectiveness of our complete pipeline}.

\evalparagraph{Missing Points}
Our method results in accurate point clouds, as seen in the high precision scores in Tab.~\ref{tab:comparison} (right).
To further evaluate the effect of missing points, we measure the recall against the synthetically removed points on \textsc{kitchen}.
We see that in a radius of 10 mm 35\% of the points are completed. 
Within 20 mm 60\% and within 30 mm 74\% of the removed points are completed.

\evalparagraph{Large-Scale Extension}
\new{We evaluate large-scale completion on the \textsc{Office} scene (73M points, 2064 images, six rooms), which is too large for any of the compared Gaussian Splatting baselines to process directly.
Using our divide-and-conquer scheme with six chunks, the scene completes in approximately 3 hours on a single A40 GPU, or roughly 70 minutes when all six chunks are processed in parallel across six GPUs.
The completed point cloud recovers thin structures throughout the multi-room scan (see \ref{supp:large_scale}).
Applying chunking to small scenes incurs a slight cost: on \textsc{Meetingroom}, a four-chunk scheme lowers F1 by approximately 0.6\%, caused by minor artifacts at chunk boundaries.}

\evalparagraph{Runtime}
\new{Our additions to baseline 2DGS (loading uncertainty maps, computing ambiguity scores, and running the filtering and sampling pipeline) account for approximately 1.2 minutes of overhead in total, as measured on the \textsc{Attic} scene.
The adapted densification and additional losses add negligible cost; the full runtime breakdown is shown in \ref{supp:runtime}.}

\evalparagraph{Pose Robustness}
\begin{figure}[]
	\centering
	\includegraphics[width=1.0\linewidth]{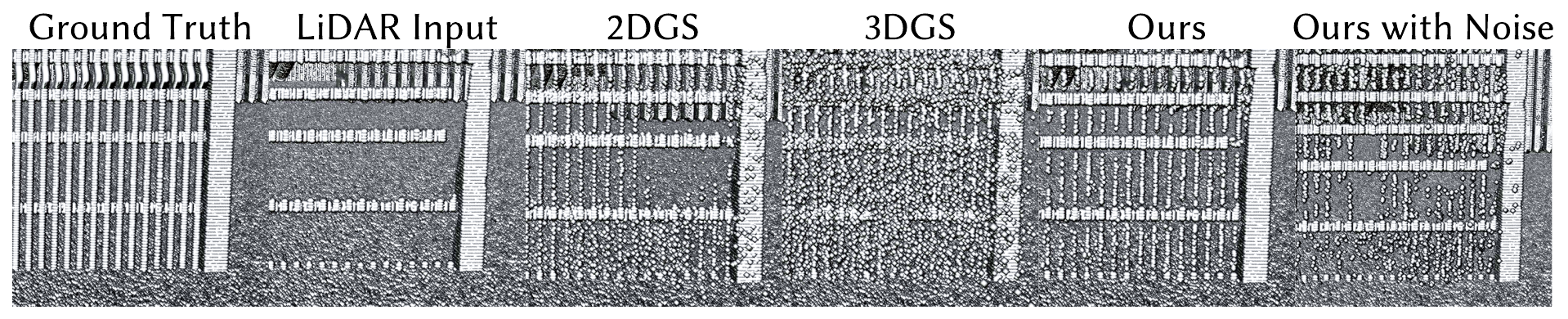}%
	\caption{Resulting point clouds for the \textsc{Attic} scene without pose noise using 2DGS, 3DGS and our method compared to our method using pose noise.}
	\label{img:nocameraerror}
\end{figure}
To evaluate the effect of pose noise to our method, we conducted an experiment on the synthetic \textsc{Attic} scene without the small pose noise added.
As seen in Fig.~\ref{img:nocameraerror}, our method is able to accurately and finely reconstruct all small structures, which is not possible with added pose noise.
Fortunately, compared to related works, our method is less susceptible to this kind of noise.

%% file: 05-conc.tex
\section{Limitation \& Future Work}

Our method has limitations inherited from photometric 3D reconstruction.
Sparse view configurations reduce reconstruction quality, particularly in areas with minimal coverage.
\new{The \textsc{Bridge} scene (350\,m, 2484 images, details in the Appendix) illustrates this: despite successfully training across 18 GPUs, the large inter-capture distance and moving objects prevent meaningful geometric recovery, though no LiDAR data is degraded.}
Dynamic elements introduce artifacts that our approach struggles to compensate. The same applies to variable lighting and view-dependent effects, which would be beneficial to address in future work.

\new{Pose noise, while tolerated to some extent, affects quality with larger deviations. Our experiments show that without pose noise, nearly all thin structures in the \textsc{Attic} scene are recovered; with the small industrial-grade noise of $0.01^\circ$ spread, some fine detail is lost. Larger deviations degrade results further.} Precise SLAM or SfM is therefore crucial. Future strategies like pose correction or incremental buildup~\cite{Fu_2024_CVPR} could alleviate this constraint.

\new{It is important to note that our method does not specifically target dark or reflective surfaces as a completion task. While we analyze these as a source of LiDAR artifacts, photometric reconstruction methods are also known to struggle in exactly these regions. Our primary contribution is the recovery of thin structures, edges, and geometric discontinuities where image cues are rich. For dark or reflective surfaces, the photometric reconstruction lacks the necessary texture signal, and our method correctly leaves these regions unchanged rather than introducing erroneous geometry. Extending the approach to these cases, for example through learned priors or multi-spectral imaging, is a direction for future work.}

\new{Residual noise around completed thin structures (e.g.\ \textsc{Kitchen} chair legs) is a known limitation. The current distance-based filter largely prevents noise from entering well-scanned flat regions, but the boundary between completed and original LiDAR geometry can still accumulate sparse outliers. Explicit post-processing denoising or a learned outlier rejection step could further improve precision in these transition zones.}

\new{More fundamentally, our completed points do not undergo the same structured post-processing as the original LiDAR data. Scanner preprocessing pipelines apply voxel-based or similar schemes to achieve uniform point spacing (typically around 5\,mm), making the original scan geometrically regular. Our points are sampled from Gaussian primitives and lack this regularity, which can make completed regions appear visually noisier than the surrounding LiDAR data. Applying equivalent structuring would require access to the scanner's proprietary preprocessing software, which is not available. Quantitative metrics such as Chamfer Distance and F1-score are therefore the most reliable basis for evaluating completion quality, as they measure geometric accuracy independently of point distribution regularity.}

\new{Finally, the non-uniform downsampling step, while essential for tractable optimization, carries an inherent risk: if a thin structure has already been almost entirely removed by the LiDAR pre-processing filter, the remaining transition-region points may be too sparse or too few to reliably guide Gaussian growth back into the correct location. In such cases, the completion may recover only a partial structure or miss it entirely. This represents a fundamental limit of our approach: we can only complete what the transition region still faintly implies, not if complete large structures have disappeared without trace. In practice, however, we never encountered this as an issue.}

\section{Conclusion}
We introduced SurfFill, a novel LiDAR point cloud completion approach using a focused Gaussian surfel splatting technique.
Our method effectively addresses gaps in LiDAR point clouds caused by common artifacts, which we analyze in this work. 
We introduce using an \textit{ambiguity heuristic} and grow surfels into missing areas, completing the scan with sampled points. 
A divide-and-conquer extension allows our method to complete point clouds in large-scale scenarios, such as building scans.
Extensive experiments on several challenging datasets verify the effectiveness of our method.

%% file: 99-suppl.tex
\section{Implementation and Training Details}
\label{supp:impl_detail}

Efficient GPU memory usage is crucial to our completion scheme. 
Thus, we cap the number of Gaussians in optimization, pausing densification when resources are fully utilized. 
To further optimize memory, we also reduce the number of spherical harmonics bands to one, using only diffuse colors.

Training is performed for 25,000 iterations, with density recalculated every 8000 epochs.
We skip the warm-up phase~\cite{kerbl20233d} and initiate densification after 100 epochs, leveraging the geometrically accurate initialization from the scan. 
Additionally, opacity is reset only every 7,500 epochs.
Uncertainty masks are precomputed using the method from Bae et al.~\cite{bae2021estimating} with ScanNet weights and preloaded as binary masks for efficient training.
Also, when computing $p(\mathbf{x})$, we scale with $f=1000$ to avoid floating point inaccuracies.
For our KNN-searches during the ambiguity heuristic and filtering, while higher $k$ theoretically yields more accurate results, using $k=3$ and $k=5$ sufficed for great results.
Especially employing the \textit{simple-knn}~\cite{kerbl20233d} framework for the former decreases processing time considerably.

For fisheye cameras commonly used with scans~\cite{yeshwanthliu2023scannetpp, cui2024letsgo}, we undistort images $\mathbf{I}_D$ into $90^{\circ}$ pinhole images, optionally with multiple crops using different rotations.
We use three crops for the wide field of view NavVis VLX captures and one for ScanNet++.

\new{\paragraph{Software environment and coordinate alignment}
Our method is implemented in Python~3.10 on top of the 2DGS codebase~\cite{huang20242d}, itself derived from the 3DGS framework~\cite{kerbl20233d}. All experiments are run with PyTorch~2.9.1 and CUDA~12.8 on Nvidia A40 GPUs (48\,GB VRAM).
The \textit{simple-knn} CUDA extension~\cite{kerbl20233d} is used in three stages: (1)~computing the ambiguity score $p(\mathbf{x})$ for each point during preprocessing, (2)~initializing Gaussian scales from nearest-neighbor distances at the start of training, and (3)~computing curvature-based scores during the filtering step. Open3D is used for point cloud I/O and for the spatial KDTree queries in the distance-based filter (Sec.~4.4 of the main paper). The monocular surface normal estimator~\cite{bae2021estimating} requires pretrained ScanNet weights, which are downloaded prior to training and used solely in the uncertainty map pre-computation step.
LiDAR point clouds and camera poses are expected to share a common coordinate frame. For our own datasets this is provided natively by the NavVis scanner software. For Tanks \& Temples and ScanNet++, we use the camera poses and LiDAR data directly as provided by the respective datasets, without additional alignment. No further registration step is required.}

\section{Hyperparameter Study}
\label{supp:hyperstudy}
\begin{table}[]
\centering
\footnotesize
\begin{minipage}[t]{0.21\textwidth}
\centering
\begin{tabular}{@{}cc@{}}
\toprule
\textbf{Point Ambiguity} & \textbf{F1-score} \\
\textbf{Threshold} & \textbf{$\uparrow$} \\
\midrule

0.02 & 0.9507 \\
\cellcolor{green!50}0.04 & \cellcolor{green!50}0.9509 \\
0.06 & 0.9501 \\
0.08 & 0.9502 \\
\bottomrule
\end{tabular}
\vspace{0.3em}
\smallskip
\end{minipage}
\hfill
\begin{minipage}[t]{0.21\textwidth}
\centering
\begin{tabular}{@{}cc@{}}
\toprule
\textbf{Uncertainty Map} & \textbf{F1-score} \\
\textbf{Threshold} & \textbf{$\uparrow$} \\
\midrule
0.0 & 0.9504 \\
0.1 & 0.9506 \\
\cellcolor{green!50}0.2 & \cellcolor{green!50}0.9509 \\
0.3 & 0.9503 \\
\bottomrule
\end{tabular}

\smallskip
\end{minipage}
\caption{Resulting F1-scores for the synthetic attic scene using varying hyperparameters: (a) Point Ambiguity Threshold and (b) Uncertainty Map Threshold. The best-performing values in each group are highlighted in green.}
\label{tab:ablation_fscore}
\end{table}
To identify optimal settings for our method, we perform an ablation study on two key hyperparameters. The \textit{Point Ambiguity Threshold} is applied during multiple stages of our pipeline: initial preprocessing, focused Gaussian splatting and filtering to discard low-ambiguity points. The \textit{Uncertainty Map Threshold} is used to selectively mask ambiguous regions in image space when applying the color, edge, and depth distortion as well as normal consistency losses from 2D Gaussian Splatting as described in the main paper. Table~\ref{tab:ablation_fscore} shows the F1-scores across different values for each parameter for the synthetic attic scene. We observe that a threshold of 0.04 for point ambiguity and 0.2 for the uncertainty mask yield the best results.

\section{Exploratory Noise Generation}
\label{supp:noise_details}

We adapt Kheradmand et al.~\cite{kheradmand20243d} in our approach for 2D surfels.
This allows for more spatial exploration along the tangential vectors with 
\begin{equation}
    \mathbf{n} = (\mathbf{L}\mathbf{L}^T) \cdot l_n \cdot \text{sigmoid}(-k(1-\alpha-t)) \cdot \mathcal{N}(0,1) . %
    \label{eq:pos_noise}
\end{equation}
Hereby, $\mathbf{L}$ describes the noise shape according to Huang et al.~\cite{huang20242d}.
During rendering, they represent 2D Gaussians as a $4\times4$ homogeneous transformation matrix $\mathbf{H}$ with 
\begin{equation}
\mathbf{H} = 
    \begin{bmatrix}
       \mathbf{L} & \mathbf{p} \\
        0 & 1 \\
    \end{bmatrix}
    \quad\text{with}\quad \mathbf{L} = 
    \begin{bmatrix}
        \mathit{s_u} \cdot \mathbf{t}_u & \mathit{s_v} \cdot \mathbf{t}_v & {0} \\
    \end{bmatrix}
    .
    \label{2dgs}
\end{equation}
$l_n$ is a scheduled hyperparameter initialized to 20. The sigmoid function with $k=100$ and $t=0.995$ describes the sharp cutoff around the opacity cutoff~\cite{kheradmand20243d}.

\section{Loss and Regularizer Experiments}
\begin{table*}[]
    \centering
    \caption{\label{tab:scenes}Overview of the scenes evaluated on.}%
    \setlength{\tabcolsep}{6pt}
    \footnotesize
    \begin{tabular}{cc|cc|ccccc}
        Dataset & Scene & LiDAR scanner & \# Points & \# Images & Captured Resolution & Used Resolution & Capturing Modality\\\hline
        Own &  \textsc{Attic} & Synthetic & 18M&708 &700x700& 700x700 & Pinhole \\
        Own &  \textsc{Kitchen} & Synthetic & 9.4M &492 &1024x1024& 1024x1024 & Pinhole \\
        Own &  \textsc{Museum} & Synthetic & 6.6M&708 &700x700& 700x700 & Pinhole \\
        Own &  \textsc{Real Attic} & NavVis VLX 3 & 17M &708& 3648x5472 &700x700 &Fisheye\\
        Own &  \textsc{Bridge} & NavVis VLX 3 & 231M & 2484 & 3648x5472 &700x700 &Fisheye\\
        Own &  \textsc{Office} & NavVis VLX & 73M & 2064 & 3648x5472 &1024x1024 &Fisheye\\
        ScanNet++ &  \textsc{1ae9e5d2a6} & FARO Focus Premium & 19M & 193 & 1752x1168 & 1168x778  &Fisheye\\
        ScanNet++ &  \textsc{45d2e33be1} & FARO Focus Premium & 16M & 226 & 1752x1168 & 1168x778 &Fisheye\\
        ScanNet++ &  \textsc{f94c225e84} & FARO Focus Premium & 33M & 426 & 1752x1168 & 1168x778 &Fisheye\\
        T\&T&  \textsc{Caterpillar} & FARO Focus 3D X330 HDR & 6.1M & 383 & 1920x1080 & 1365x768 &Pinhole\\ 
        T\&T&  \textsc{Meetingroom} & FARO Focus 3D X330 HDR & 43M & 371 & 1920x1080 & 1365x768 &Pinhole\\  
        T\&T&  \textsc{Courthouse} & FARO Focus 3D X330 HDR & 59M & 1106 & 1920x1080 & 1365x768 &Pinhole\\  
        T\&T&  \textsc{Truck} & FARO Focus 3D X330 HDR & 8M & 251 & 1920x1080 & 1365x768 &Pinhole 
    \end{tabular}
    \vspace{-3mm}
\end{table*}

Accurate loss formulation proved crucial for our approach. 
We implemented several adjustments to Gaussian Splatting that did not improve geometric quality and were discarded. 
Inspired by previous work, we explored the incorporation of depth or normal priors for regularization. 
Using LiDAR-measured depth was ineffective as the missing structures were absent in the depth maps. 
Alternatively, we tested normal and depth maps estimated from visual data. 
Normals estimated using the method by Bae et al.~\cite{bae2021estimating} produced significant errors in high-curvature regions, discouraging splats in critical areas and degrading results. 
Similarly, depth maps estimated with Depth Anything~\cite{yang2024depth} resulted in poor geometric outcomes.

We also experimented with using the perceptual VGG loss~\cite{johnsonvgg16} instead of the L1 loss for penalizing visual errors. 
The VGG loss, compares feature vectors of target and ground truth images, capturing perceptual patterns. 
However, during Gaussian Splatting training, it introduced significant artifacts. For example, noise and motion blur from the input images became amplified.
The loss penalized general patterns, like noise, rather than our regions of interest, resulting in no improvement in Gaussian positions. 
Consequently, this approach was discarded.

\section{Test Scene Details}
\label{supp:test_scene_details}

\begin{figure}[]
	\centering
	\includegraphics[width=1.0\linewidth]{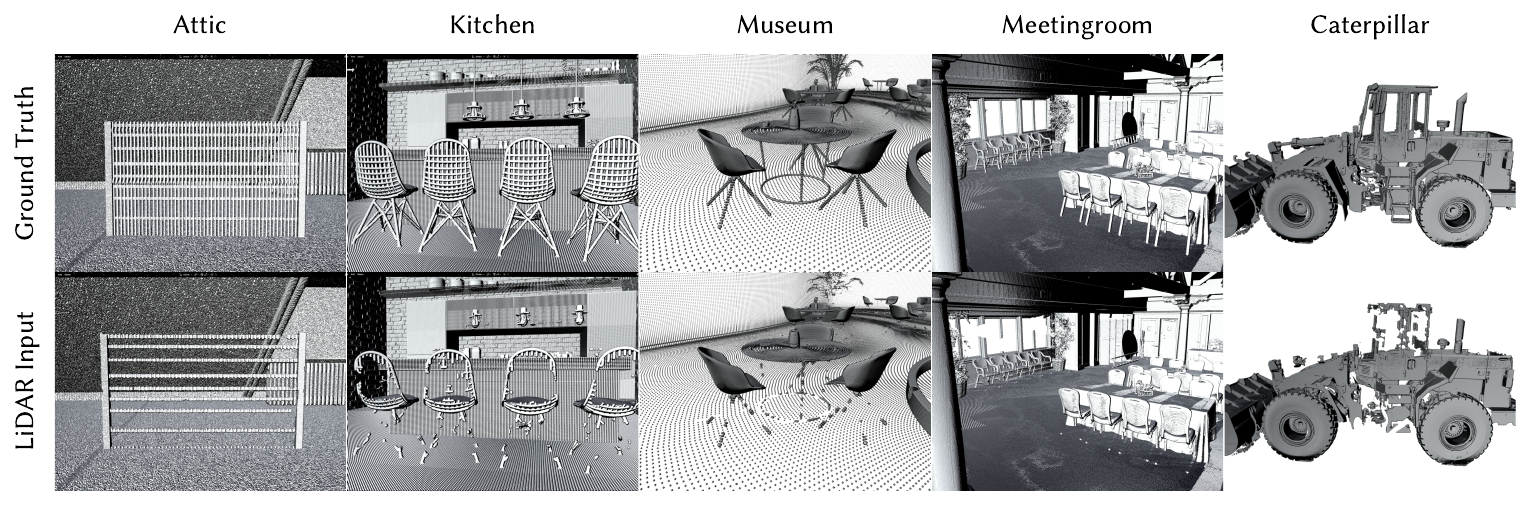}
    \caption{Ground truth and constructed LiDAR input point clouds for our test scenes.}
	\label{img:comparison_grid_lidardeleting}
\end{figure}

For synthetic evaluation against ground truth, we use five test scenarios. 
Fig.~\ref{img:comparison_grid_lidardeleting} shows the areas we delete from the ground truth point clouds to generate the input "LiDAR" data. For this, we mimic the behavior of a LiDAR sensor. For the \textsc{Attic} scene, which has a real-world counterpart, we delete the exact regions missing in the real dataset. 

Table~\ref{tab:scenes} provides an overview of the scene statistics for our five test scenes and five real-world datasets.

\section{Additional Evaluations}
\label{supp:additional_evals}
In this section, we present the results of additional experiments conducted for our method. 

\subsection{Expanded F1-scores}
\begin{table}[]
\centering
\caption{\label{tab:comparison_full}Full F1-Score results for our test scenes enhanced using different LiDAR completion methods: Our method achieves the highest performance, followed by using 3DGS.}
\footnotesize
\setlength{\tabcolsep}{2pt}
\begin{tabular}{l|ccccc|c}
Completion & \multicolumn{6}{c}{F1-score $\uparrow$} \\
\text{Method} & \textsc{  Attic  } & \textsc{  Kitchen  } & \textsc{ Museum } & \textsc{Meetingr.} & \textsc{Caterpillar} & \text{ Mean }  \\\hline
\text{Neuralangelo}  & 0.9016 & 0.7247 & 0.7942 & 0.8294 & 0.6328 & 0.7765  \\
\text{COLMAP}  & 0.9357 & 0.6962 & 0.8640 & 0.9408 & 0.6338 & 0.8141   \\
\text{3DGS}  & 0.9256 & 0.7958 & 0.8318 & 0.9717 & 0.8689 & \cellcolor{yellow}0.8795\\
\text{2DGS} & \cellcolor{yellow}0.9383 & 0.6993 & \cellcolor{yellow}0.8715 & \cellcolor{yellow}0.9804 & \cellcolor{yellow}0.9012 & 0.8781  \\
\text{GOF}  & 0.9382 & \cellcolor{yellow}0.8191 & 0.8217 & 0.9614 & 0.8252 & 0.8731 \\
\text{Ours} &  \cellcolor{green!50}0.9509 & \cellcolor{green!50}0.8419 & \cellcolor{green!50}0.8855 & \cellcolor{green!50}0.9877 & \cellcolor{green!50}0.9223 & \cellcolor{green!50}0.9176  \\
\end{tabular}
\end{table}

Table \ref{tab:comparison_full} presents the complete set of F1-scores computed for all test scenes, providing a more detailed evaluation than the mean F1-scores and Chamfer Distances reported in the main paper. These results further demonstrate that our method consistently outperforms all other compared approaches across the full range of test cases.

\subsection{Large-scale scenes}
\label{supp:large_scale}
\begin{figure}[]
	\centering
	\includegraphics[width=0.99\linewidth]{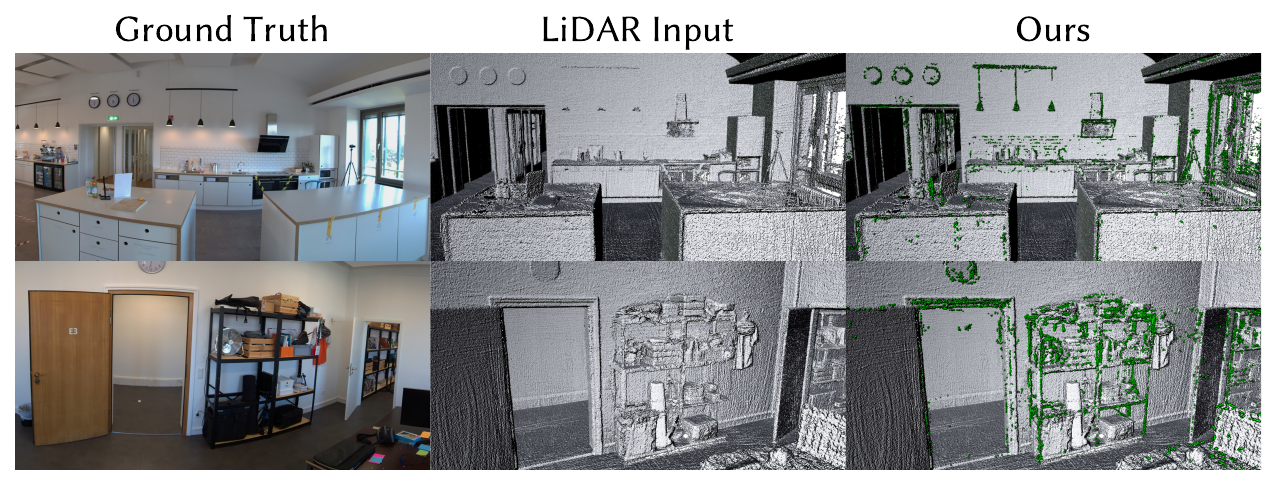}
    \vspace{-3mm}%
	\caption{Results of the improved point cloud of the large real-world LiDAR \textsc{Office} dataset trained with our method using 6 chunks  computed in roughly 72 minutes. Points added by our algorithm are highlighted in green. }
	\label{img:comparison_grid_largeNavvisRoom}
    \vspace{-3mm}
\end{figure}
Our method aims at completing LiDAR point clouds in large-scale scenarios, for which we employ a divide-and-conquer scheme. %
To test this scheme, we use a scan of a multi-room office building, where six rooms were captured.
This scan encompasses 73M points as well as 2064 images, and as such is not processable by related Gaussian Splatting works.

We can process it in six chunks and complete the point cloud, as seen in Fig.~\ref{img:comparison_grid_largeNavvisRoom}.
Hereby, the total processing time is around 3 hours on a single A40 GPU and about 70 minutes on six A40 GPUs.

However, we note that applying chunking to small-scale scenes does not improve results.
For example on the \textsc{Meetingroom} scene, using a four-chunk scheme lowers F1-scores by about 0.6\%, as slight artifacts at chunk edges decrease precision scores.
This could be addressed in future work by a smarter blending of chunks.

\begin{figure*}[]
	\centering
	\includegraphics[width=1.0\linewidth]{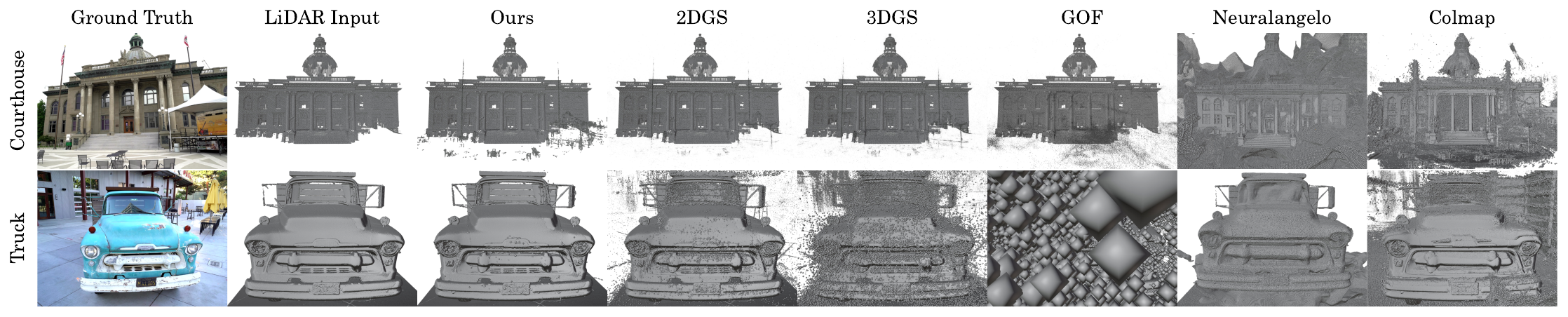}
    \vspace{-5mm}
	\caption{Resulting point clouds for the real-world datasets \textsc{courthouse} and \textsc{truck} from Tanks \& Temples trained with the evaluated methods.}
	\label{img:tnttruck}
\end{figure*}

\subsection{Direct Point Cloud Completion Comparison}
\label{supp:direct_pc_comp}
\begin{figure}[]
	\centering
	\includegraphics[width=1.0\linewidth]{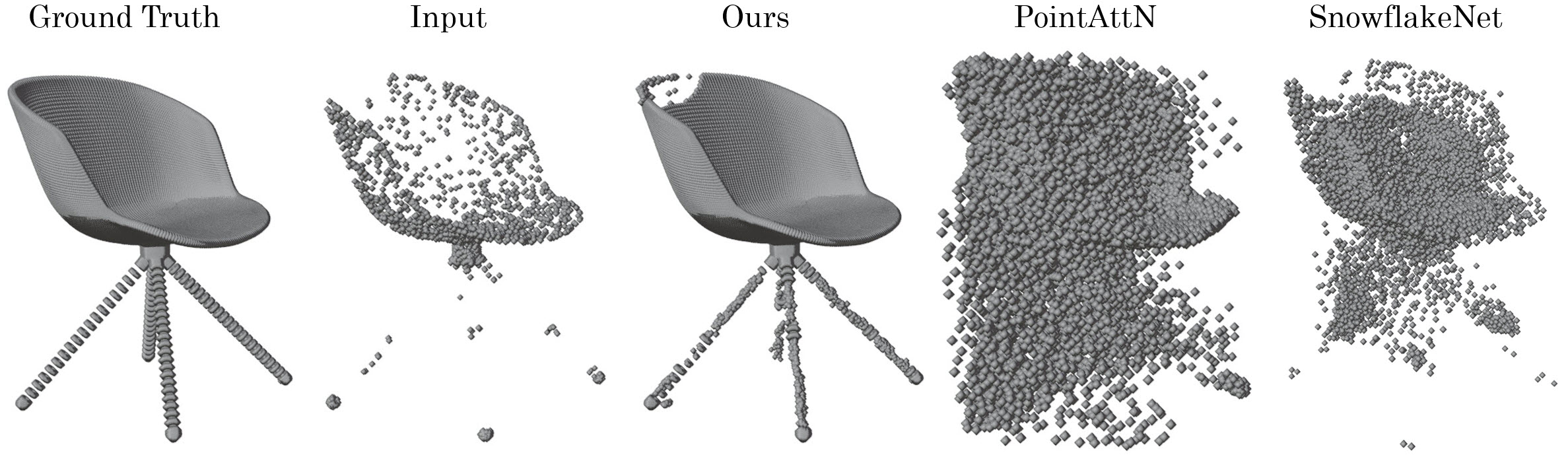}
	\caption{Comparison between our method, SnowflakeNet~\cite{xiang2021snowflakenet} and PointAttN~\cite{pointattn} for the simple chair example.}
	\label{img:pointattnchair}
\end{figure}

To evaluate the effectiveness of our method in completing complex LiDAR point clouds, we compare it to the state-of-the-art direct point cloud completion methods PointAttN~\cite{pointattn} and SnowflakeNet~\cite{xiang2021snowflakenet}. PointAttN and SnowflakeNet are designed to complete small partial point clouds using deep learning. Current popular point cloud completion methods mainly revolve around the design of an encoder-decoder architecture for complete point cloud generation. SnowflakeNet emphasizes the decoding process by introducing a skip-transformer to model spatial relationships across multiple decoding stages. While its attention mechanism effectively captures structural features in point clouds, it still relies on k-nearest neighbors (kNN) to model local geometric relationships. In contrast, PointAttN avoids explicit local region partitioning like kNN, making it more robust to variations in point cloud density. Instead, it leverages cross-attention and self-attention mechanisms to establish both short- and long-range relationships among points using two key modules: Geometric Details Perception (GDP) and Self-Feature Augment (SFA)~\cite{pointattn}. 

We selected PointAttN and SnowflakeNet for comparison because they currently achieve some of the best results on the Completion3D~\cite{completion3d} and PCN datasets~\cite{pcndataset}, which are often used to benchmark direct point cloud completion methods. The datasets include single-object point clouds across various categories, including chairs. For our experiment, we test reconstruction performance on two simple examples extracted from our synthetic museum scene: a single chair and a table accompanied by two chairs. To test the effectiveness of our method, we train our museum scene as described in the evaluation section of the main paper. We then extract the completed chair and table \& chair point clouds from the entire trained scene and compare them to their corresponding synthetic ground truth point clouds. Since PointAttN and SnowflakeNet operate on 2048-point inputs, we first extract partial synthetic LiDAR point clouds for the chair and chair-table set from the museum scene to test these methods. We then uniformly downsample the point clouds to 2048 points and complete them using the PointAttN and SnowflakeNet models trained on the PCN dataset. Finally, we compare the point clouds completed with the PointAttN and SnowflakeNet method to their ground truth counterparts. The results, shown in Figure~\ref{img:pointattnchair} and Table~\ref{tab:pointattn}, reveal that PointAttN and SnowflakeNet perform significantly worse than our approach. SnowflakeNet achieves an average F1-score of just 0.1054 compared to 0.8918 for our method. PointAttN achieves an even lower score. Despite having been trained on similar objects, PointAttN and SnowflakeNet appear to generalize poorly to similar point clouds with different characteristics, such as those from LiDAR. This highlights a key limitation of most direct point cloud completion approaches: they are constrained to small-scale, clean input data and struggle when applied to real-world scenes with high point counts and complex structures.

\begin{table}[]
\centering
\caption{\label{tab:pointattn}Quantitative comparison between our method and the direct point cloud completion methods PointAttN~\cite{pointattn} and SnowflakeNet~\cite{xiang2021snowflakenet}.}
\renewcommand{\arraystretch}{1.1} %
\setlength{\tabcolsep}{6pt} %
\footnotesize
\begin{tabular}{l|c|c|c|c}
Completion & Large & \multicolumn{2}{c|}{Chamfer Distance (m) $\downarrow$} & F1-score $\uparrow$ \\

\text{Method} & Scale & Chair & Chairs \& Table & Mean \\
\hline
\text{PointAttN} & no & 0.0554 & \cellcolor{yellow}0.1274 & 0.0858 \\
\text{SnowflakeNet} & no & \cellcolor{yellow}0.0413 & 0.1576 & \cellcolor{yellow}0.1054 \\
\text{Ours} & yes & \cellcolor{green!50}0.0011 & \cellcolor{green!50}0.0177 & \cellcolor{green!50}0.8918 \\
\end{tabular}
\end{table}

\subsection{Runtime Inspection}
\label{supp:runtime}
\begin{figure}[]
	\centering
	\includegraphics[width=1.0\linewidth]{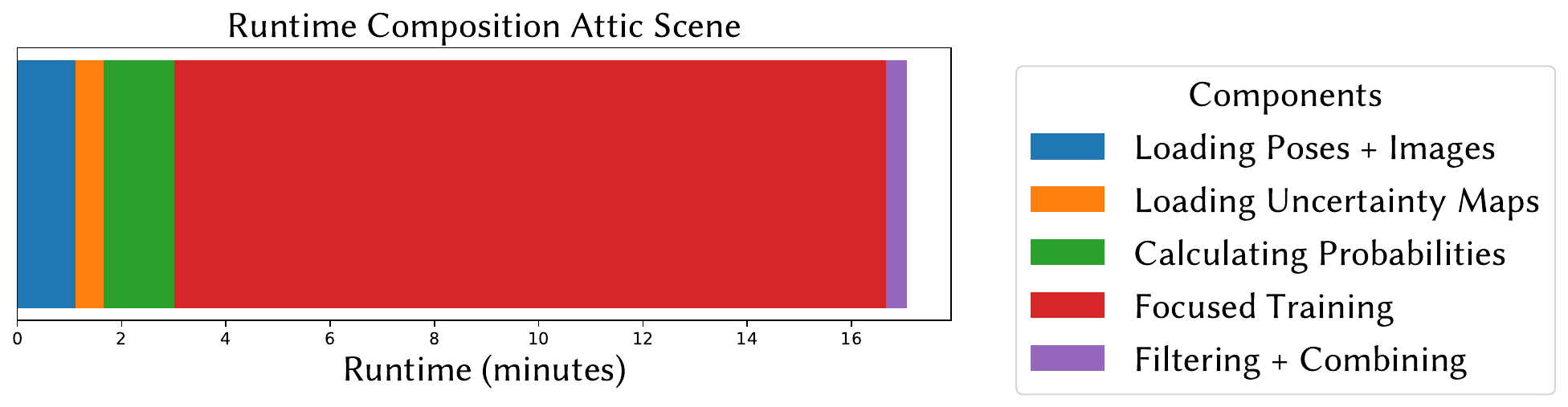}
	\caption{Computation time of each step of our method for the \textsc{Attic} scene.}
	\label{img:runtime_composition_training}
\end{figure}

In Fig.~\ref{img:runtime_composition_training}, we present the runtime composition measured for the \textsc{Attic} scene. Loading the uncertainty maps, calculating the probabilities and filtering and combining the final point clouds are steps not present in baseline 2DGS. The runtime of these stages roughly sums up to the total average difference in runtime of 1.2 minutes between the two methods. This shows that the added losses and adapted densification scheme of our focused surfel splatting method have almost no effect on our runtime performance.

\subsection{Further Tests on Tanks \& Temples}
\label{supp:tnt2}

In Fig.~\ref{img:tnttruck}, additional qualitative real-world results are shown for the \textsc{Truck} and \textsc{Courthouse} datasets from Tanks \& Temples. 
Our method successfully reconstructs the missing thin beam in the \textsc{Truck} dataset without introducing noticeable noise. 
While most other methods also reconstruct this structure, they add significantly more outliers. 
The \textsc{courthouse} scene suffers from missing structures surrounding the building, such as the flag posts, chairs and tent. 
Our method once again reconstructs the most missing elements without adding a noticeable amount of noise. 
Our added points also integrate best with the original LiDAR data due to our surfel to point sampling scheme.

\subsection{Chunking Artifacts}
\label{supp:chunking_artifacts}
\begin{figure}[]
	\centering
	\includegraphics[width=1.0\linewidth]{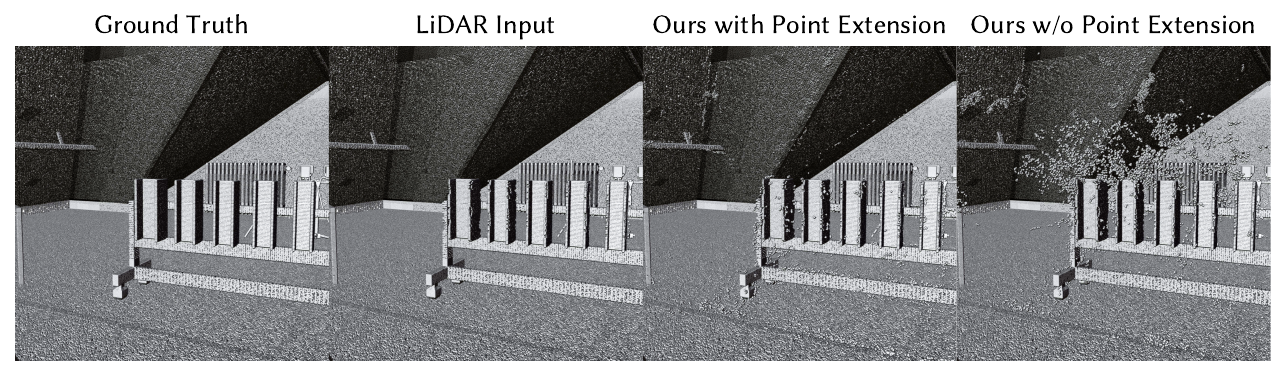}
    \vspace{-6mm}
	\caption{Resulting Point Clouds for the \textsc{Attic} Test Scene using our method with 2x2 chunks
 with and w/o Point Extension: point extension reduces artifacts at the chunks’ borders.}
	\label{img:pointextension}
\end{figure}

We show the effect of the \textit{point extension step} necessary for dataset chunking in Fig.~\ref{img:pointextension}. 
For this experiment, the \textsc{Attic} scene is subdivided into 2 $\times$ 2 chunks. 
In the ”Point Extension” case the point clouds for each chunk are generated as specified by the algorithm of Kerbl et al.~\cite{kerbl2024hierarchical} and Lin et al.~\cite{lin2024vastgaussian}. 
If point extension is left out, the chunk datasets only consist of point clouds containing the points located inside each chunk’s bounding box. 
In this case, the algorithm attempts to reduce the large visual error caused by the missing initialization for areas observed by the cameras outside the chunk. 
As a result, floating artifacts are generated. 
Due to extending the point clouds, we generally observe smooth transitions between neighboring chunks in real world datasets for our LiDAR completion application.

\begin{figure}[]
	\centering
	\includegraphics[width=1.0\linewidth]{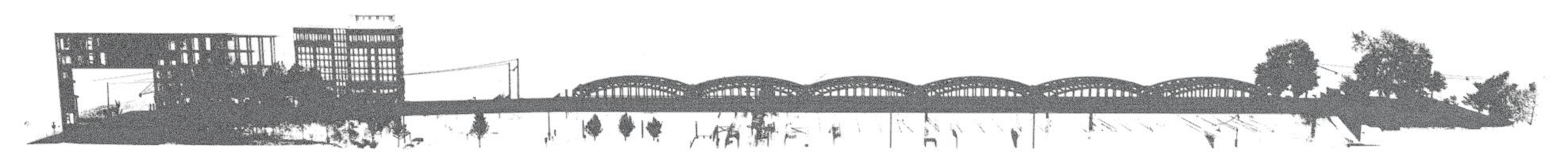}    
    \vspace{-6mm}
	\caption{Massive 350m long \textsc{Bridge} dataset scanned with a NavVis VLX 3 with 2484 images. We train this scene with 18 chunks in 2.63 hours in parallel on 18 Nvidia A40 GPUs.}
	\label{img:hackerbrueckeoverview}
    \vspace{-2mm}
\end{figure}

\subsection{Further Large-scale Tests}
\label{supp:bridge}
\begin{figure}[]
	\centering
	\includegraphics[width=1.0\linewidth]{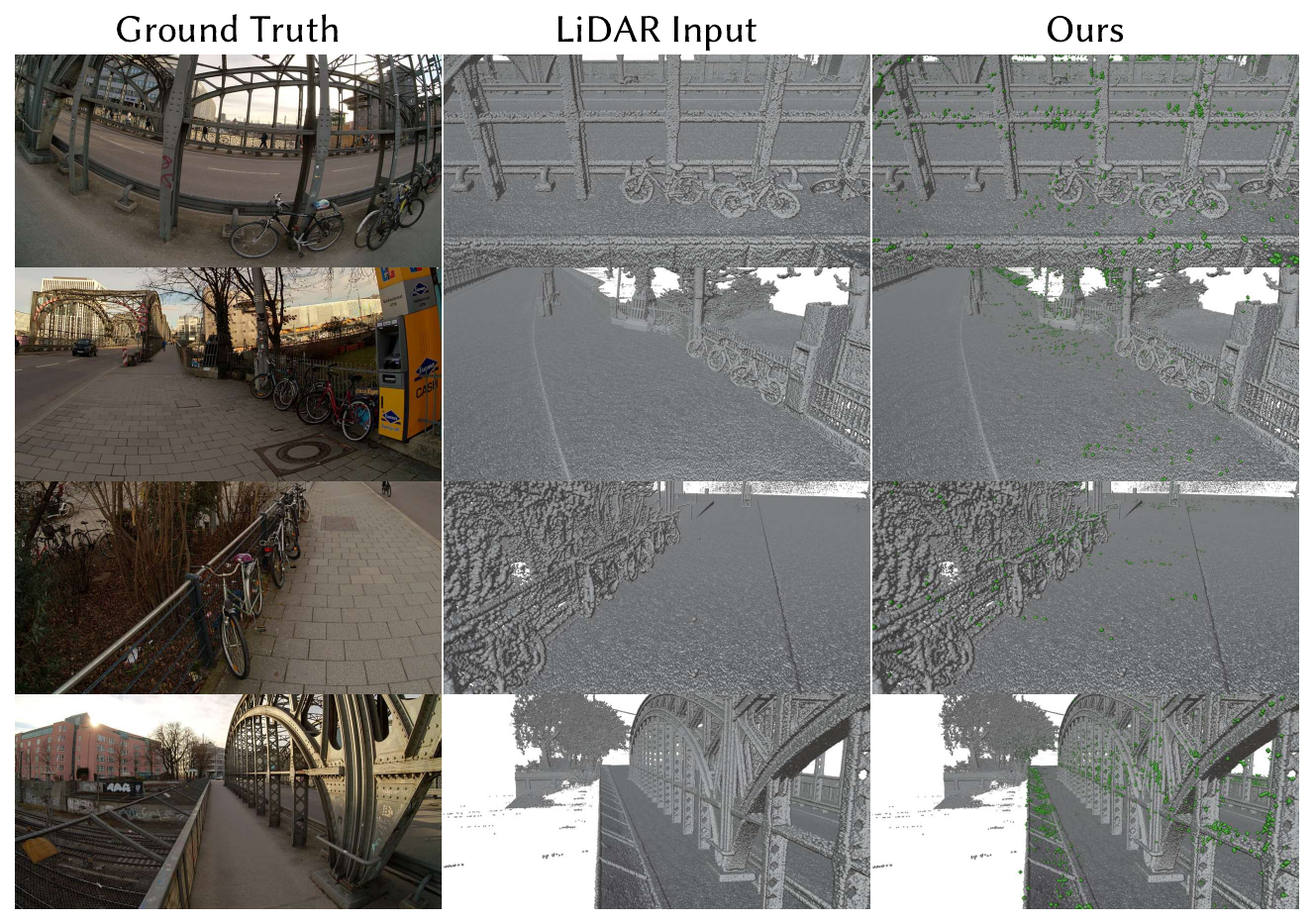}
    \vspace{-6mm}
	\caption{Resulting views of the improved point cloud of the massive real-world LiDAR \textsc{Bridge} dataset trained with our method using 18 chunks: Points added by our algorithm are highlighted in green. Our method fails to impactfully improve the LiDAR data due to the poor visual data. However, we successfully train the 350m long scene with 2484 images in 2.63h. }
    \vspace{-3mm}
	\label{img:hackerbruecke}
\end{figure}

We report the results of training a massive \textsc{Bridge} dataset with our method that has a very sparse image dataset in Fig.~\ref{img:hackerbruecke}. 
The input LiDAR point cloud can be seen in Fig.~\ref{img:hackerbrueckeoverview}. 
The RGB images show many dynamic elements, such as moving vehicles, people and variable lighting conditions. 
The dataset covers a 350m long scene with 2484 images and 231M points. 
Although we successfully train the dataset in 2.63h using 18  Nvidia A40 GPUs, we fall short of achieving the necessary geometric fidelity to recover missing structures. However, we do not worsen the LiDAR data. This highlights the limitations of our method. 
Our performance heavily depends on the quality and consistency of the captured visual data, as well as pose accuracy. 
In this dataset, very sparse images are captured with large distances between capture points, while moving geometry slightly degrades pose estimation.

\subsection{Visual Metrics}

\begin{table}[]
\caption{Resulting visual quality in terms of PSNR for our test scenes.}
\centering
\renewcommand{\arraystretch}{1.1} %
\setlength{\tabcolsep}{5pt} %
\footnotesize
\begin{tabular}{l|ccccc}
\text{Method} & \textsc{Attic} & \textsc{Kitchen} & \textsc{Museum} & \textsc{Meetingroom} & \textsc{Caterpillar} \\
\hline
\text{2DGS} & \cellcolor{green!50}25.26 & 20.18 & \cellcolor{yellow}20.45 & \cellcolor{yellow}22.09 & \cellcolor{yellow}19.02 \\
\text{3DGS} & \cellcolor{yellow}25.01 & 20.34 & 19.85 & \cellcolor{green!50}22.43 & \cellcolor{green!50}20.04 \\
\text{GOF}  & 24.39 & \cellcolor{green!50}21.05 & 18.25 & 16.03 & 13.34 \\
\text{Ours} & 24.99 & \cellcolor{yellow}20.37 & \cellcolor{green!50}21.32 & 20.77 & 16.11 \\
\end{tabular}
\label{tab:psnr_comparison}
\end{table}

\begin{table}[]
\centering
\caption{Visual performance comparison of different Gaussian Splatting methods on synthetic test scenes and the  \textsc{Meetingroom} and \textsc{Caterpillar} test scenes from Tanks and Temples.}
\footnotesize
\setlength{\tabcolsep}{5pt}
\renewcommand{\arraystretch}{1.1}
\begin{tabular}{l|ccc|ccc}
\textbf{} & \multicolumn{3}{c|}{\textbf{Synthetic Scenes}} & \multicolumn{3}{c}{\textbf{Tanks and Temples}} \\
\text{Method} & \text{PSNR $\uparrow$} & \text{SSIM $\uparrow$} & \text{LPIPS $\downarrow$} & \text{PSNR $\uparrow$} & \text{SSIM $\uparrow$} & \text{LPIPS $\downarrow$} \\
\hline
2DGS          & \cellcolor{yellow}21.96 & \cellcolor{yellow}0.578 & 0.427 & \cellcolor{yellow}20.55 & \cellcolor{green!50}0.659 & \cellcolor{yellow}0.317 \\
3DGS & 21.73 & 0.575 & \cellcolor{yellow}0.411 & \cellcolor{green!50}21.24 & \cellcolor{yellow}0.658 & \cellcolor{green!50}0.265 \\
GOF   & 21.23 & 0.567 & 0.435 & 14.68 & 0.519 & 0.501 \\
Ours          & \cellcolor{green!50}22.10 & \cellcolor{green!50}0.590 & \cellcolor{green!50}0.390 & 18.37 & 0.611 & 0.354 \\
\end{tabular}
\label{tab:performance_comparison}
\end{table}
While we do not focus on visual reconstruction quality, we test if our method achieves visual results comparable to other state-of-the-art Gaussian Splatting methods. 
For this, we create visual test datasets for our synthetic scenes by rendering additional views and use 25\% of the captured images of the TnT datasets for testing. We exclude these views from the training process. We report the resulting PSNR, SSIM and perceptual LPIPS losses in Tables~\ref{tab:psnr_comparison} and~\ref{tab:performance_comparison}. 
We find that our method does not significantly degrade visual quality, but also falls short of visually outperforming the other methods.